\documentclass{article}

\usepackage[nonatbib,preprint]{neurips_2026}

\usepackage[utf8]{inputenc}
\usepackage[T1]{fontenc}
\usepackage{url}
\usepackage{booktabs}
\usepackage{amsfonts}
\usepackage{microtype}
\usepackage[table]{xcolor}
\usepackage{listings}
\usepackage{algorithm}
\usepackage{algorithmic}
\usepackage{wrapfig}
\usepackage{array}
\usepackage{abbrv}
\usepackage{enumitem}
\usepackage{xspace}
\usepackage{xparse}
\usepackage{graphicx, amsmath, amssymb, caption, subcaption, multirow}
\usepackage{tikz}
\usetikzlibrary{shapes.geometric, arrows.meta}
\usepackage{pgfplots}
\pgfplotsset{compat=1.18}

\definecolor{accentorange}{RGB}{217,119,87}
\definecolor{axisgray}{RGB}{192,185,175}
\definecolor{slateblue}{RGB}{90,105,120}
\definecolor{modelgreen}{RGB}{120,140,110}
\definecolor{reflinecolor}{RGB}{180,120,100}
\definecolor{softsand}{RGB}{245,240,235}

\definecolor{citecolor}{HTML}{0071BC}
\definecolor{linkcolor}{HTML}{ED1C24}
\definecolor{codebasecolor}{HTML}{2A6F97}
\usepackage[pagebackref=false, breaklinks=true, letterpaper=true, colorlinks, bookmarks=false,  citecolor=citecolor, linkcolor=linkcolor, urlcolor=gray]{hyperref}
\makeatletter\renewcommand\paragraph{\@startsection{paragraph}{4}{\z@}
  {.5em \@plus1ex \@minus.2ex}{-.5em}{\normalfont\normalsize\bfseries}}\makeatother

\newlength\savewidth\newcommand\shline{\noalign{\global\savewidth\arrayrulewidth
  \global\arrayrulewidth 1pt}\hline\noalign{\global\arrayrulewidth\savewidth}}
\newcommand{\tablestyle}[2]{\setlength{\tabcolsep}{#1}\renewcommand{\arraystretch}{#2}\centering\footnotesize}

  \newcolumntype{x}[1]{>{\centering\arraybackslash}p{#1pt}}
\newcolumntype{y}[1]{>{\raggedright\arraybackslash}p{#1pt}}
\newcolumntype{z}[1]{>{\raggedleft\arraybackslash}p{#1pt}}

\newcommand{\app}{\raise.17ex\hbox{$\scriptstyle\sim$}}

\definecolor{deemph}{gray}{0.7}
\newcommand{\gc}[1]{\textcolor{gray}{#1}}
\definecolor{baselinecolor}{RGB}{236,232,225}
\newcommand{\baseline}[1]{\cellcolor{baselinecolor}{#1}}
\newcommand{\baselinelegend}[1]{%
  \begingroup
  \setlength{\fboxsep}{1pt}%
  \colorbox{baselinecolor}{\raisebox{0pt}[1.2ex][.2ex]{#1}}%
  \endgroup
}
\newcommand{\authorskip}{\hspace{2.5mm}}

\newcommand{\method}{{\emph{FD-loss}}\xspace}

\definecolor{basebg}{RGB}{236,232,225}
\definecolor{oursbg}{RGB}{250,230,216}

\definecolor{Ncol}{RGB}{210,230,255}
\definecolor{Bcol}{RGB}{255,218,200}
\DeclareRobustCommand{\mathhl}[2]{%
  \tikz[baseline=(X.base)] \node[fill=#1, rounded corners=1pt, inner sep=1pt] (X)
  {$\displaystyle #2$};}
\DeclareRobustCommand{\hlN}[1]{\mathhl{Ncol}{#1}}
\DeclareRobustCommand{\hlB}[1]{\mathhl{Bcol}{#1}}
\newcommand{\frechet}{{Fr\'echet}\xspace}

\NewDocumentCommand{\fdr}{g}{%
  \ensuremath{\text{FDr}\IfValueT{#1}{^{#1}}}\xspace
}
\newcommand{\apprange}[2]{pp.~\pageref{#1}--\pageref{#2}}

\title{Representation \frechet Loss for Visual Generation}

\author{%
  Jiawei Yang\textsuperscript{1}\authorskip
  Zhengyang Geng\textsuperscript{2}\authorskip
  Xuan Ju\textsuperscript{3}\authorskip
  Yonglong Tian\textsuperscript{4}\authorskip
  Yue Wang\textsuperscript{1}\\[0.5em]
  \textsuperscript{1}USC\quad
  \textsuperscript{2}CMU\quad
  \textsuperscript{3}CUHK\quad
  \textsuperscript{4}OpenAI
}

\begin{document}

\maketitle

\input{sections/figures/teaser_figure}

\begin{abstract}
We show that \frechet Distance (FD), long considered impractical as a training objective, can in fact be effectively optimized in the representation space.
Our idea is simple: decouple the \emph{population size} for FD estimation (\eg, 50k) from the \emph{batch size} for gradient computation (\eg, 1024). 
We term this approach \method.
Optimizing \method reveals several surprising findings. 
First, post-training a base generator with \method in different representation spaces consistently improves visual quality. Under the Inception feature space, a one-step generator achieves \emph{0.72} FID on ImageNet $256\times256$.
Second, the same \method repurposes multi-step generators into strong one-step generators \emph{without} teacher distillation, adversarial training or per-sample targets. 
Third, FID can misrank visual quality: modern representations can yield better samples despite worse Inception FID. This motivates \fdr{k}, a multi-representation metric.
We hope this work will encourage further exploration of distributional distances in diverse representation spaces as both training objectives and evaluation metrics for generative models.
Code and checkpoints are available at {\hypersetup{urlcolor=codebasecolor}\href{https://github.com/Jiawei-Yang/FD-loss}{https://github.com/Jiawei-Yang/FD-loss}}.
\end{abstract}

\section{Introduction}
\label{sec:intro}

\definecolor{figdark}{HTML}{141413}        %
\definecolor{figmid}{HTML}{B0AEA5}         %
\definecolor{figlightgray}{HTML}{E8E6DC}   %
\definecolor{figorange}{HTML}{D97757}      %
\definecolor{figblue}{HTML}{6A9BCC}        %
\definecolor{figdarkbrown}{HTML}{3D3229}   %
\definecolor{figcream}{HTML}{F5F0EB}       %
\definecolor{figfdtext}{RGB}{105,66,65}    %
\definecolor{figfdfill}{RGB}{244,238,238}  %
\definecolor{figfdborder}{RGB}{197,179,178} %
\definecolor{gradred}{HTML}{CC3333}        %
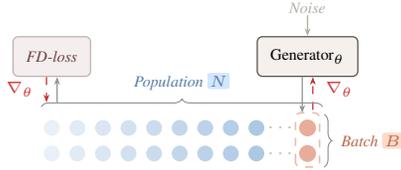
\begin{figure}[t]
\centering
\vspace{-1em}
\begin{minipage}{0.44\linewidth}{
\vspace{-1.1em}
\centering
\begin{tikzpicture}[
    >={Stealth[length=3pt]},
    box/.style={draw=figdark!70, rounded corners=2pt, minimum height=1.5em,
                inner xsep=5pt, inner ysep=2pt,
                font=\tiny, line width=0.5pt},
    lbl/.style={font=\tiny, text=figdark},
    brace/.style={decorate, decoration={brace, amplitude=3pt}, figdark!50},
    bracemirror/.style={decorate, decoration={brace, amplitude=3pt, mirror}, figdark!50},
  ]

  \def\dx{0.34}
  \def\cx{1.53}  %

  \node[box, fill=figcream] (gen) at ({10*\dx}, 2.15)
    {Generator$_\theta$};
  \node[lbl, text=figmid, font=\tiny\itshape] at ({10*\dx}, 2.8) {Noise};
  \draw[->, line width=0.5pt, figmid] ({10*\dx}, 2.7) -- (gen.north);

  \draw[brace] ({10*\dx+0.22}, 1.38) -- ({10*\dx+0.22}, 0.68)
    node[midway, right=3pt, lbl, text=figorange!85!black]
    {\textit{Batch}~\hlB{B}};

  \foreach \i in {0,...,8} {
    \pgfmathsetmacro{\opacity}{15 + \i * 5}
    \fill[figblue!\opacity] (\i*\dx, 1.2) circle (3pt);
  }
  \node[font=\tiny, figmid] at ({9*\dx}, 1.2) {$\cdots$};
  \fill[figorange!60] ({10*\dx}, 1.2) circle (3.2pt);
  \foreach \i in {0,...,8} {
    \pgfmathsetmacro{\opacity}{15 + \i * 5}
    \fill[figblue!\opacity] (\i*\dx, 0.86) circle (3pt);
  }
  \node[font=\tiny, figmid] at ({9*\dx}, 0.86) {$\cdots$};
  \fill[figorange!60] ({10*\dx}, 0.86) circle (3.2pt);

  \draw[figorange!50, dashed, rounded corners=2pt, line width=0.6pt]
    ({10*\dx-0.16}, 0.68) rectangle ({10*\dx+0.16}, 1.4);

  \draw[->, line width=0.5pt, figdark!45] ([xshift=-2pt]gen.south) -- ({10*\dx-0.07}, 1.4);
  \draw[->, line width=0.5pt, gradred, dashed] ({10*\dx+0.07}, 1.44) -- ([xshift=2pt]gen.south);
  \node[lbl, text=gradred, anchor=west, font=\tiny] at ({10*\dx+0.12}, 1.72)
    {$\nabla_\theta$};

  \draw[brace] (-0.14, 1.48) -- ({10*\dx+0.2}, 1.48)
    node[midway, above=3pt, lbl, text=figblue!80!black]
    {\textit{Population}~\hlN{N}};

  \node[draw=figfdborder, fill=figfdfill, rounded corners=2pt,
        line width=0.5pt, minimum height=1.5em,
        inner xsep=5pt, inner ysep=2pt,
        font=\tiny, text=figfdtext] (fd) at (0, 2.15)
    {\method};

  \draw[->, line width=0.5pt, figdark!45] ([xshift=2pt]fd.south |- {0,1.54}) -- ([xshift=2pt]fd.south);
  \draw[->, line width=0.5pt, gradred, dashed] ([xshift=-2pt]fd.south) -- ([xshift=-2pt]fd.south |- {0,1.54});
  \node[lbl, text=gradred, anchor=east, font=\tiny] at (-0.12, 1.72)
    {$\nabla_\theta$};

\end{tikzpicture}
}\end{minipage}
\hfill
\begin{minipage}{0.53\linewidth}{
\caption{\textbf{\method.} The generator produces \hlB{B} images per step; their features are accumulated into a large population via a queue or EMA. This decouples the population size \hlN{N} for reliable FD estimation from the batch size \hlB{B} for gradient computation.}
\label{fig:decouple}
}\end{minipage}
\vspace{-1em}
\end{figure}

The \frechet Inception Distance (FID) \cite{heusel2017gans} has been the \emph{de facto} metric for evaluating image generation, especially on academic benchmarks such as ImageNet~\cite{deng2009imagenet}. For nearly a decade, the community has been collectively performing ``gradient descent'' on this single metric to push the state of the art. But this ``gradient descent'' has always been \emph{indirect}: FID serves only as an evaluator, \emph{not} as a loss. 
In this paper, we ask: \emph{can FID be optimized directly as a training loss---and what does that reveal?}

In principle, this is straightforward: every term in \frechet Distance (FD) is differentiable and nothing in its definition restricts it to evaluation. In practice, however, this has been widely considered impractical. Computing FID requires a large population of samples (\eg, 50k) to estimate distributional statistics. Using it as a training loss would further demand gradients through all of them at every step, which is prohibitive. Prior work has therefore only explored FD as a loss under restricted settings, \eg, estimating FD from a small batch~\cite{mathiasen2020backpropagating,doan2020image}. Our experiments confirm that optimizing batch-wise FD degrades base generators (Tab.~\ref{tab:queue_size}).

In this work, we show that FD can in fact be optimized directly \emph{at scale}. Our idea is simple (Fig.~\ref{fig:decouple}): \emph{decouple} the \emph{population size} used for FD estimation from the \emph{batch size} used for gradient computation. We call this approach \method and realize it in two ways. The first maintains an online queue of features from recently generated samples and computes FD over the full queue (\eg, 50k) while back-propagating only through the current batch. The second maintains exponential moving averages (EMA) of the first and second moments of features and restricts gradients to the current batch.
Both work well in practice.
This one approach unlocks several \emph{surprising} findings.

First, \method is a strong post-training objective for visual generators.
We fine-tune a pre-trained generator with \method using only pre-computed feature statistics---means and covariances of real data under one or more representation backbones.\footnote{Real data is used only once, offline, to compute these reference statistics. No real images are seen during post-training.}
Across pixel-space and latent-space generator families~\cite{lu2026one,geng2025improved,li2025back}, model sizes, and image resolutions, this consistently improves visual quality.
Under Inception~\cite{szegedy2016rethinking}, \method drives a \emph{one-step} generator~\cite{lu2026one} to an FID of {0.72} on ImageNet $256\times256$ (Tab.~\ref{tab:system}). The same recipe, applied under modern representations~\cite{liu2022convnet,he2022masked,oquab2024dinov2,radford2021learning,tschannen2025siglip}, yields even stronger perceptual quality (Fig.~\ref{fig:repr_single}, Tab.~\ref{tab:all_fd_models}).

Second, \method serves as a simple distribution-matching objective.
We show that a pre-trained \emph{multi-step} generator can be repurposed into a \emph{one-step} generator by post-training it with \method (from around 300 FID to 0.72). This repurposing works without teacher distillation, adversarial training, or per-sample targets, suggesting that the representation-space FD is a useful distribution-level objective rather than only an evaluation metric.
We validate this on both class-conditioned models (Fig.~\ref{fig:teaser_samples}, bottom) and text-conditioned models (Fig.~\ref{fig:t2i}).

Third, \method provides a diagnostic lens. 
Having models trained by \method with different representations, we can test whether the best-FID model is also the perceptually best. 
Often, it is \emph{not}: models optimized with modern representations achieve better visual quality while scoring \emph{worse} under FID (Tab.~\ref{tab:backbone}, Fig.~\ref{fig:repr_single}). More broadly, this connects to a paradox in the field (Fig.~\ref{fig:scatter_benchmark}): state-of-the-art generators already surpass \emph{real} validation images under FID, yet their outputs remain clearly distinguishable from real images. 
We therefore report \fdr{k}, a normalized FD ratio averaged over $k$ feature spaces (Eq.~\ref{eq:fdrk}), as a more representation-diverse automatic metric. As shown in Figure~\ref{fig:scatter_benchmark}, FID suggests the ``ImageNet generation'' problem is nearly solved, yet \fdr{k} reveals that significant quality gaps remain. Post-training with \method narrows this gap, achieving an \fdr{6} of {1.89}.

\providecommand{\ltiny}{\fontsize{5pt}{6pt}\selectfont}
\providecommand{\ttiny}{\fontsize{4pt}{5pt}\selectfont}
\providecommand{\tttiny}{\fontsize{3pt}{4pt}\selectfont}
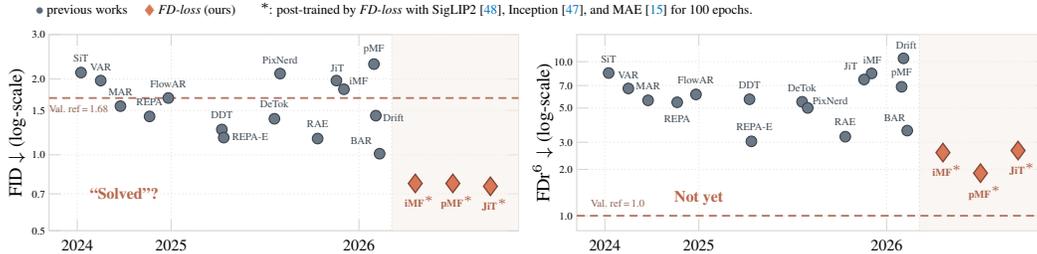
\begin{figure}[t]
\begin{center}
\hspace{-.8em}
\begin{tikzpicture}[remember picture]
\begin{axis}[
  name=fidplot,
  width=0.56\textwidth,
  height=0.3\textwidth,
  xlabel={},
  ylabel={FID $\downarrow$ (log-scale)},
  ylabel style={at={(-0.03,0.5)}, anchor=south},
  xmin=2023.7, xmax=2028.7,
  ymin=0.5, ymax=3,
  ymode=log,
  log basis y=10,
  xtick={2024, 2025, 2027},
  xticklabels={2024, 2025, 2026},
  xticklabel style={/pgf/number format/1000 sep={}},
  ytick={0.5, 0.7, 1, 1.5, 2, 3},
  yticklabels={0.5, 0.7, 1.0, 1.5, 2.0, 3.0},
  x tick label style={font=\tiny},
  y tick label style={font=\ttiny, xshift=2pt},
  major grid style={axisgray!40, densely dotted, line width=0.4pt},
  minor grid style={axisgray!20, dotted, line width=0.3pt},
  grid=major,
  label style={font=\scriptsize},
  axis line style={axisgray!60, line width=0.4pt},
  tick style={axisgray!60, line width=0.3pt},
  scatter/classes={
    sd={mark=*, mark size=2pt, draw=slateblue!60!black, fill=slateblue!90},
    nosd={mark=*, mark size=2pt, draw=slateblue!60!black, fill=slateblue!90}
  },
]
\fill[softsand, opacity=0.6] (axis cs:2027.35, 0.5) rectangle (axis cs:2028.7, 3);

\draw[reflinecolor, densely dashed, line width=0.7pt] (axis cs:2023.7, 1.68) -- (axis cs:2028.7, 1.68);
\node[font=\ttiny, reflinecolor!80!black, anchor=south west] at (axis cs:2023.6, 1.35) {Val.\ ref\,=\,1.68};

\draw[axisgray!40, densely dotted, line width=0.5pt] (axis cs:2027.35, 0.5) -- (axis cs:2027.35, 3);

\addplot[scatter, only marks, scatter src=explicit symbolic]
coordinates {
  (2024.77, 1.42) [sd]  %
  (2025.54, 1.26) [sd]   %
  (2025.56, 1.17) [sd]   %
  (2026.56, 1.16) [sd]   %
  (2027.22, 1.01) [sd]   %
};
\addplot[scatter, only marks, scatter src=explicit symbolic]
coordinates {
  (2024.04, 2.12) [nosd]  %
  (2024.25, 1.97) [nosd]  %
  (2024.97, 1.68) [nosd]  %
  (2024.46, 1.56) [nosd]  %
  (2026.10, 1.39) [nosd]  %
  (2026.84, 1.82) [nosd]  %
  (2026.76, 1.97) [nosd]  %
  (2026.16, 2.10) [nosd]  %
  (2027.18, 1.43) [nosd]  %
  (2027.16, 2.29) [nosd]  %
};

\addplot[only marks, mark=diamond*, mark size=3.5pt,
  fill=accentorange, draw=accentorange!60!black, line width=0.5pt]
coordinates {
  (2027.6, 0.77)   %
  (2028.0, 0.77)   %
  (2028.4, 0.75)   %
};

\node[font=\ttiny, anchor=south, slateblue!70!black] at (axis cs:2024.77, 1.45) {REPA};
\node[font=\ttiny, anchor=south, slateblue!70!black] at (axis cs:2025.54, 1.28) {DDT};
\node[font=\ttiny, anchor=south west, slateblue!70!black] at (axis cs:2025.55, 1.05) {REPA-E};
\node[font=\ttiny, anchor=south, slateblue!70!black] at (axis cs:2026.56, 1.18) {RAE};
\node[font=\ttiny, anchor=south east, slateblue!70!black] at (axis cs:2027.24, 1.01) {BAR};
\node[font=\ttiny, anchor=south, slateblue!70!black] at (axis cs:2024.04, 2.15) {SiT};
\node[font=\ttiny, anchor=south, slateblue!70!black] at (axis cs:2024.25, 1.99) {VAR};
\node[font=\ttiny, anchor=south, slateblue!70!black] at (axis cs:2024.97, 1.70) {FlowAR};
\node[font=\ttiny, anchor=south, slateblue!70!black] at (axis cs:2024.46, 1.58) {MAR};
\node[font=\ttiny, anchor=south, slateblue!70!black] at (axis cs:2026.10, 1.41) {DeTok};
\node[font=\ttiny, anchor=south west, slateblue!70!black] at (axis cs:2026.80, 1.74) {iMF};
\node[font=\ttiny, anchor=south west, slateblue!70!black] at (axis cs:2026.6, 1.97) {JiT};
\node[font=\ttiny, anchor=south, slateblue!70!black] at (axis cs:2026.16, 2.12) {PixNerd};
\node[font=\ttiny, anchor=south west, slateblue!70!black] at (axis cs:2027.16, 1.25) {Drift};
\node[font=\ttiny, anchor=south, slateblue!70!black] at (axis cs:2027.16, 2.31) {pMF};
\node[font=\ttiny\bfseries, anchor=north, accentorange!80!black] at (axis cs:2027.65, 0.75) {iMF$^*$};
\node[font=\ttiny\bfseries, anchor=north, accentorange!80!black] at (axis cs:2028.05, 0.75) {pMF$^*$};
\node[font=\ttiny\bfseries, anchor=north, accentorange!80!black] at (axis cs:2028.45, 0.73) {JiT$^*$};

\node[font=\tiny\bfseries, accentorange!85!black, align=center] at (axis cs:2024.5, 0.7) {``Solved''?};

\end{axis}
\hspace{0.2em}
\begin{axis}[
  name=fdrplot,
  at={(fidplot.east)},
  anchor=west,
  xshift=20pt,
  width=0.56\textwidth,
  height=0.3\textwidth,
  xlabel={},
  ylabel={\fdr{6} $\downarrow$ (log-scale)},
  ylabel style={at={(-0.03,0.5)}, anchor=south},
  xmin=2023.7, xmax=2028.7,
  ymin=0.8, ymax=15,
  ymode=log,
  log basis y=10,
  xtick={2024, 2025, 2027},
  xticklabels={2024, 2025, 2026},
  xticklabel style={/pgf/number format/1000 sep={}},
  x tick label style={font=\tiny},
  y tick label style={font=\ttiny, xshift=2pt},
  ytick={1, 2, 3, 5, 7, 10},
  yticklabels={1.0, 2.0, 3.0, 5.0, 7.0, 10.0},
  major grid style={axisgray!40, densely dotted, line width=0.4pt},
  minor grid style={axisgray!20, dotted, line width=0.3pt},
  grid=major,
  label style={font=\scriptsize},
  axis line style={axisgray!60, line width=0.4pt},
  tick style={axisgray!60, line width=0.3pt},
  scatter/classes={
    sd={mark=*, mark size=2pt, draw=slateblue!60!black, fill=slateblue!90},
    nosd={mark=*, mark size=2pt, draw=slateblue!60!black, fill=slateblue!90}
  },
]
\fill[softsand, opacity=0.6] (axis cs:2027.35, 0.8) rectangle (axis cs:2028.7, 15);

\draw[reflinecolor, densely dashed, line width=0.7pt] (axis cs:2023.7, 1.0) -- (axis cs:2028.7, 1.0);
\node[font=\ttiny, reflinecolor!80!black, anchor=south west] at (axis cs:2023.75, 1.0) {Val.\ ref\,=\,1.0};

\draw[axisgray!40, densely dotted, line width=0.5pt] (axis cs:2027.35, 0.8) -- (axis cs:2027.35, 15);

\addplot[scatter, only marks, scatter src=explicit symbolic]
coordinates {
  (2024.77, 5.45) [sd]  %
  (2025.54, 5.70) [sd]   %
  (2025.56, 3.04) [sd]   %
  (2026.56, 3.26) [sd]   %
  (2027.22, 3.57) [sd]   %
};
\addplot[scatter, only marks, scatter src=explicit symbolic]
coordinates {
  (2024.04, 8.44) [nosd]  %
  (2024.25, 6.70) [nosd]  %
  (2024.97, 6.13) [nosd]  %
  (2024.46, 5.61) [nosd]  %
  (2026.10, 5.49) [nosd]  %
  (2026.84, 8.39) [nosd]  %
  (2026.76, 7.66) [nosd]  %
  (2026.16, 5.01) [nosd]  %
  (2027.18, 10.51) [nosd]  %
  (2027.16, 6.87) [nosd]  %
};

\addplot[only marks, mark=diamond*, mark size=3.5pt,
  fill=accentorange, draw=accentorange!60!black, line width=0.5pt]
coordinates {
  (2027.6, 2.57)   %
  (2028.0, 1.89)   %
  (2028.4, 2.65)   %
};

\node[font=\ttiny, anchor=south, slateblue!70!black] at (axis cs:2024.77, 3.5) {REPA};
\node[font=\ttiny, anchor=south, slateblue!70!black] at (axis cs:2025.54, 5.80) {DDT};
\node[font=\ttiny, anchor=south west, slateblue!70!black] at (axis cs:2025.3, 3.14) {REPA-E};
\node[font=\ttiny, anchor=south, slateblue!70!black] at (axis cs:2026.56, 3.36) {RAE};
\node[font=\ttiny, anchor=south east, slateblue!70!black] at (axis cs:2027.3, 3.57) {BAR};
\node[font=\ttiny, anchor=south, slateblue!70!black] at (axis cs:2024.04, 8.54) {SiT};
\node[font=\ttiny, anchor=south, slateblue!70!black] at (axis cs:2024.25, 6.80) {VAR};
\node[font=\ttiny, anchor=south, slateblue!70!black] at (axis cs:2024.97, 6.23) {FlowAR};
\node[font=\ttiny, anchor=south, slateblue!70!black] at (axis cs:2024.46, 5.71) {MAR};
\node[font=\ttiny, anchor=south, slateblue!70!black] at (axis cs:2026.10, 5.59) {DeTok};
\node[font=\ttiny, anchor=south west, slateblue!70!black] at (axis cs:2026.65, 8.49) {iMF};
\node[font=\ttiny, anchor=south west, slateblue!70!black] at (axis cs:2026.45, 7.76) {JiT};
\node[font=\ttiny, anchor=south, slateblue!70!black] at (axis cs:2026.40, 4.6) {PixNerd};
\node[font=\ttiny, anchor=south west, slateblue!70!black] at (axis cs:2027.0, 10.61) {Drift};
\node[font=\ttiny, anchor=south, slateblue!70!black] at (axis cs:2027.16, 6.8) {pMF};
\node[font=\ttiny\bfseries, anchor=north, accentorange!80!black] at (axis cs:2027.65, 2.47) {iMF$^*$};
\node[font=\ttiny\bfseries, anchor=north, accentorange!80!black] at (axis cs:2028.05, 1.79) {pMF$^*$};
\node[font=\ttiny\bfseries, anchor=north, accentorange!80!black] at (axis cs:2028.45, 2.55) {JiT$^*$};

\node[font=\tiny\bfseries, accentorange!85!black, align=center] at (axis cs:2025.0, 1.3) {Not yet};

\end{axis}
\node (legend) [anchor=south west, inner sep=2.5pt,
  rounded corners=1.5pt, font=\ltiny] at ([xshift=-10pt, yshift=4pt]fidplot.north west) {%
  {\color{slateblue}$\bullet$}\,previous works\hspace{6pt}%
  {\color{accentorange}$\blacklozenge$}\,\method (ours)\hspace{8pt}
  $^*$\!: post-trained by \method with SigLIP2~\cite{tschannen2025siglip}, Inception~\cite{szegedy2016rethinking}, and MAE~\cite{he2022masked} for 100 epochs.%
};
\end{tikzpicture}
\end{center}

\vspace{-0.75em}
\caption{
\textbf{Is ImageNet generation ``solved''?}
\textbf{Left}: FID over time. Recent methods surpass the \emph{real} validation set images (red dashed) in terms of FID. \textbf{Right}: \fdr{6} (Eq.~\ref{eq:fdrk}), which averages normalized \frechet Distance ratios across six representation spaces. Under this metric, even the strongest existing methods remain far from the validation images, indicating that FID alone masks significant quality gaps. Each method is shown at its largest publicly available model size. Post-training with \method ({\color{accentorange}$\blacklozenge$}) improves both metrics; see Table~\ref{tab:system} for numerical results.
}
\label{fig:scatter_benchmark}
\end{figure}

Taken together, our findings reposition FD in generative modeling. FD has long lived on one side of the generative modeling pipeline, \ie, \emph{evaluation}. We show it is also useful on the other side, \ie, \emph{training}. The two sides turn out to be linked: making FD as a loss both improves generators and exposes the limits of any single FD as an evaluator. We hope the simplicity of \method will encourage the community to rethink both how we train and how we evaluate generative models.

\section{Related Work}
\label{sec:related_work}

\paragraph{\frechet Distance as an evaluation metric.}
FID~\cite{heusel2017gans} is the dominant metric for evaluating image generation. A growing body of work has highlighted its limitations~\cite{sajjadi2018assessing,kynkaanniemi2019improved,kynkaanniemi2022role,stein2023exposing,jayasumana2024rethinking}, motivating alternatives such as precision--recall~\cite{sajjadi2018assessing,kynkaanniemi2019improved}, CKA~\cite{yang2023revisiting}, and MMD~\cite{jayasumana2024rethinking}. We make FD directly optimizable as a training loss, which both probes the reliability of FID and motivates our \fdr{k} metric.

\paragraph{\frechet Distance as a training loss.}
Distributional distances as training objectives date back to adversarial learning~\cite{goodfellow2014generative}, MMD-based generators~\cite{li2015generative,binkowski2018demystifying,zhou2025inductive,deng2026generative}, and sliced Wasserstein objectives~\cite{deshpande2018generative}. More closely related, several works train generators by matching feature-space moments~\cite{mroueh2017mcgan,santos2019learning}, or by minimizing FD in Inception~\cite{mathiasen2020backpropagating} or discriminator~\cite{doan2020image} feature space. The main limitation of existing FD optimization work is statistical scale: their FD estimates are computed within a single batch and become too noisy to scale up. Our work makes FD optimization practical \emph{at scale}.

\paragraph{Optimizing over broader sample windows.}
A recurring difficulty in modern deep learning optimization is that some objectives require a much larger effective sample set than a single batch provides. In contrastive learning, this motivates memory banks~\cite{wu2018unsupervised} and feature queues~\cite{he2020momentum}. Deep networks have also long used exponential moving averages (EMA) to maintain stable estimates of population statistics, \eg, Batch Normalization~\cite{ioffe2015batch}. Our work adopts a similar principle: we compute FD over a queue of recent features or EMA estimates of feature moments, while back-propagating only through the current batch.

\paragraph{One-step generators and post-training.}
Modern high-quality image generators often rely on multi-step denoising~\cite{Peebles2023,esser2024scaling,li2025autoregressive,zheng2025diffusion,mo2025group}. This has motivated more efficient one-step or few-step generators, including straightening ODEs~\cite{liu2023flowstraight}, consistency models~\cite{song2023consistency,lu2025simplifying}, score distillation~\cite{luo2023diffinstruct,luo2024scoreimplicit,yin2024one,zhou2024score}, identity-based methods~\cite{geng2025mean,geng2025improved,lu2026one}, and drifting models~\cite{deng2026generative}. Our work is complementary: optimizing FD in capable representation spaces is a powerful and new way to improve existing one-step generators~\cite{lu2026one,geng2025improved} and to repurpose multi-step generators~\cite{li2025back,esser2024scaling} into one-step ones, in both pixel and latent space, \emph{without} denoising teacher distillation or adversarial training. This positions FD not only as an evaluation tool, but also as a practical post-training objective.

\section{Method}
\label{sec:method}

We study using \frechet Distance (FD) as a training objective for post-training image generators. We begin by outlining the challenges of directly optimizing FD, and then present our method for making this practical at scale.

\subsection{Preliminaries: \frechet Distance}
\label{sec:fd}

Let $\phi(\cdot)$ denote a feature extractor. Given real images $\mathcal{R}=\{\mathbf{x}_i\}$ and generated images $\mathcal{G}=\{\hat{\mathbf{x}}_i\}$, their feature distributions are modeled as multivariate Gaussians with means and covariances:
\begin{equation}
  \boldsymbol{\mu}_r = \mathbb{E}[\phi(\mathbf{x})], \quad
  \boldsymbol{\Sigma}_r = \mathrm{Cov}[\phi(\mathbf{x})],
  \qquad
  \boldsymbol{\mu}_g = \mathbb{E}[\phi(\hat{\mathbf{x}})], \quad
  \boldsymbol{\Sigma}_g = \mathrm{Cov}[\phi(\hat{\mathbf{x}})].
\end{equation}
The FD between the two Gaussian distributions is:
\begin{equation}
  \mathrm{FD}_{\phi}(\mathcal{R}, \mathcal{G})
  =
  \|\boldsymbol{\mu}_r - \boldsymbol{\mu}_g\|_2^2
  +
  \mathrm{Tr}\!\Big(
  \boldsymbol{\Sigma}_r + \boldsymbol{\Sigma}_g
  - 2(\boldsymbol{\Sigma}_r \boldsymbol{\Sigma}_g)^{\frac{1}{2}}
  \Big).
  \label{eq:fd}
  \end{equation}
When $\phi$ is Inception-v3~\cite{szegedy2016rethinking}, this becomes the Fr\'echet Inception Distance (FID)~\cite{heusel2017gans}.

In standard evaluation, $(\boldsymbol{\mu}_r, \boldsymbol{\Sigma}_r)$ are pre-computed once from the training set, while $(\boldsymbol{\mu}_g, \boldsymbol{\Sigma}_g)$ are estimated from a large population of generated samples, typically on the order of tens of thousands.

\paragraph{Challenges.} Unlike sample-wise losses, FD is a \emph{distributional} quantity. Directly optimizing FD is difficult because the population size needed for reliable estimation (\eg, 50k) far exceeds a typical training batch (\eg, 64 to 1024). Small-batch estimates are unstable, especially for high-dimensional features. For example, reliably estimating a full-rank covariance matrix ($\boldsymbol{\Sigma}_g \in \mathbb{R}^{2048 \times 2048}$) for Inception~\cite{szegedy2016rethinking} requires far more than 2048 samples. At the same time, back-propagating through a full evaluation-sized population at every training step is computationally infeasible for most practical setups.\footnote{Admittedly, this challenge would likely vanish if one could use a batch size of 100,000, which is rarely practical but not entirely impossible.} The central problem is therefore to estimate FD using a large effective population while keeping the cost of optimization at the scale of an affordable training batch.

\subsection{\method: Decoupling Population Scale from Optimization Scale}
\label{sec:decouple}

Our key idea is decoupling: we estimate FD using statistics aggregated over a much broader sample window than the current batch, while computing gradients with respect to the current batch alone (Fig.~\ref{fig:decouple}). We consider two implementations, introduced below and summarized in Algorithm~\ref{alg:rdm}.


\definecolor{codeblue}{rgb}{0.25,0.5,0.5}
\definecolor{codekw}{rgb}{0.85, 0.18, 0.50}
\definecolor{codesign}{RGB}{0, 0, 255}
\definecolor{codefunc}{rgb}{0.85, 0.18, 0.50}

\lstdefinelanguage{PythonFuncColor}{
language=Python,
keywordstyle=\color{blue}\bfseries,
commentstyle=\color{codeblue},
stringstyle=\color{orange},
showstringspaces=false,
basicstyle=\ttfamily\small,
literate=
  {*}{{\color{codesign}* }}{1}
  {-}{{\color{codesign}- }}{1}
  {+}{{\color{codesign}+ }}{1}
  {G}{{\color{codefunc}G}}{1}
  {phi}{{\color{codefunc}phi}}{3}
  {batch_moments}{{\color{codefunc}batch\_moments}}{13}
  {all_gather}{{\color{codefunc}all\_gather}}{13}
  {compute_stats}{{\color{codefunc}compute\_stats}}{13}
  {FD}{{\color{codefunc}FD}}{2}
  {backward}{{\color{codefunc}backward}}{8}
  {step}{{\color{codefunc}step}}{4}
  {detach}{{\color{codefunc}detach}}{6}
}

\lstset{
language=PythonFuncColor,
backgroundcolor=\color{white},
basicstyle=\fontsize{7.2pt}{8.2pt}\ttfamily\selectfont,
columns=fullflexible,
breaklines=true,
captionpos=b,
}

\begin{wrapfigure}{r}{0.47\linewidth}
\vspace{-2.2em}
\centering
\begin{minipage}{0.98\linewidth}

\begin{algorithm}[H]
\caption{Post-Training with \method.}
\label{alg:rdm}
\begin{lstlisting}
# G: generator
# phi: frozen representation model
# (mu_r, sig_r): real feature statistics
# (mu_ema, M_ema): EMA mean and 2nd moment
# beta: EMA decay
# z: current batch of noise

x = G(z)
feat = phi(x)
feat = all_gather(feat) # gather across devices

# Queue version:
# gen_feats = cat([queue.detach(), feat])
# mu_g, sig_g = compute_stats(gen_feats)

# EMA version:
mu_b, M_b = batch_moments(feat)
mu_g = beta * mu_ema.detach() + (1 - beta) * mu_b
M_g  = beta * M_ema.detach()  + (1 - beta) * M_b
sig_g = M_g - mu_g @ mu_g.T

loss = FD((mu_g, sig_g), (mu_r, sig_r))
loss.backward()
optimizer.step()

# Queue version:
# queue.enqueue_and_dequeue(feat.detach())

# EMA version
mu_ema = mu_g.detach()
M_ema  = M_g.detach()
\end{lstlisting}
\end{algorithm}

\end{minipage}
\vspace{-2.5em}
\end{wrapfigure}

\paragraph{Queue-based estimator.}
Let $N$ denote the queue size, which determines the effective population used for statistics estimation (\eg, $N=100,000$). At each training iteration, the generator produces a batch of $B$ images, where $B \ll N$ (\eg, $B=1024$). We extract their features using a representation model $\phi$ and enqueue them, while removing the oldest $B$ features. FD is computed using the empirical mean and covariance of the full queue. During back-propagation, only the current batch features carry gradients; queued features from previous iterations are treated as constants. This is similar in spirit to the queue in MoCo~\cite{he2020momentum}. The dynamically updated queue makes the feature statistics \emph{on-policy}. This implementation already enables us to achieve \textbf{0.89} FID with 50 epochs post-training using a 118M-sized model on ImageNet $256\times256$ (Tab.~\ref{tab:queue_size}).

\paragraph{EMA-based estimator.} 
We also consider an approach that avoids storing feature queues entirely. Concretely, we maintain exponential moving averages (EMA) of the first and second feature moments. Let $\beta \in (0, 1)$ denote the EMA decay rate, and $\boldsymbol{\mu}^{(t)}_{\mathrm{g}}$ and $\mathbf{M}^{(t)}_{\mathrm{g}}$ denote the running estimates of the first and second moment at iteration $t$. Given the features of images generated in the current batch $\{\phi(\hat{\mathbf{x}}_i)\}_{i=1}^{B}$, we define the batch moments as:
\begin{equation}
    \boldsymbol{\mu}^{(t)}_{\mathrm{batch}}
    = \frac{1}{B}\sum_{i=1}^{B} \phi(\hat{\mathbf{x}}_i),
    \qquad        
    \mathbf{M}^{(t)}_{\mathrm{batch}}
    = \frac{1}{B}\sum_{i=1}^{B} \phi(\hat{\mathbf{x}}_i)\,\phi(\hat{\mathbf{x}}_i)^{\top},
\end{equation}
and update the running estimates as:
\begin{equation}
  \boldsymbol{\mu}^{(t)}_{g}
  =
  \beta \boldsymbol{\mu}^{(t-1)}_{g}
  +
  (1-\beta)\boldsymbol{\mu}^{(t)}_{\mathrm{batch}},
  \qquad
  \mathbf{M}^{(t)}_{g}
  =
  \beta \mathbf{M}^{(t-1)}_{g}
  +
  (1-\beta)\mathbf{M}^{(t)}_{\mathrm{batch}},
\end{equation}
and recover the covariance from
\begin{equation}
  \boldsymbol{\Sigma}^{(t)}_{g}
  =
  \mathbf{M}^{(t)}_{g}
  -
  \boldsymbol{\mu}^{(t)}_{g}
  \boldsymbol{\mu}^{(t)\top}_{g}.
\end{equation}
FD is then computed using $(\boldsymbol{\mu}^{(t)}_{g},\, \boldsymbol{\Sigma}^{(t)}_{g})$ via Equation~\ref{eq:fd}, with gradients back-propagated only through the current batch. Unlike the queue, this variant stores \emph{no} feature buffer, making it more scalable when using multiple representation models. It also provides a more on-policy estimate since EMA naturally upweights recent samples. This enables the base model to achieve \textbf{0.81} FID under the same post-training procedure (Tab.~\ref{tab:ema_beta}).

\paragraph{Discussion.}
Both variants implement the same \emph{decoupling} principle (\S\ref{sec:decouple}). The queue uses an explicit window controlled by $N$; the EMA uses a smoothed estimate controlled by $\beta$. Both work well in practice; each trades off population size against how on-policy the statistics remain.  

\paragraph{Multi-representation \method.}
\label{sec:multi_repr}

Our \method naturally supports minimizing FD measured in different representation spaces. In practice, FD can vary by orders of magnitude across representation models.
To combine losses from multiple representations $\{\phi_i\}$, we normalize each term:
\begin{equation}
  \mathcal{L}
  =
  {\textstyle\sum_i} w_i \cdot \mathcal{L}_{\phi_i},
  \qquad
  \mathcal{L}_{\phi_{i}}
  =
  \frac{\mathrm{FD}_{\phi_i}(\mathcal{R},\, \mathcal{G})}
  {\texttt{sg}\!\left(\mathrm{FD}_{\phi_i}(\mathcal{R},\, \mathcal{G})\right) + c},
  \label{eq:multi}                                  
\end{equation}
where $\texttt{sg}(\cdot)$ denotes stop-gradient, $c$ is a small constant for numerical stability, and $w_i$ are per-representation weights. This makes each term unit-scale regardless of the feature space. For simplicity, we also do normalization even when there is only one representation model. In this work, we simply use equal weights with $w_i=1$.
 
\subsection{Training setup}
\label{sec:training_setup}

We apply our \method for post-training. In all cases, we start from a pre-trained generator, called the base model, and fine-tune it with the \method.

\paragraph{Post-training one-step generators.}
Our primary setting is post-training existing one-step generators with \method. In practice, it is desirable to improve sample quality while preserving fast generation. We therefore study both pixel-space one-step generators, such as pixel-MeanFlow (pMF)~\cite{lu2026one}, and latent-space one-step generators, such as improved-MeanFlow (iMF)~\cite{geng2025improved}, to demonstrate the versatility of our method.

\paragraph{Repurposing multi-step generators.}
Surprisingly, we find that the \method, when measured in capable representation spaces, can repurpose a pre-trained multi-step generator into a one-step generator. Concretely, given Gaussian noise $z$, we run the model only once at the terminal timestep and interpret its output as a one-step prediction of the clean image. For example, $\hat{x}_0 = z - v_\theta(z,\; t{=}1)$ for a velocity-prediction model (\eg, SiT~\cite{ma2024sit} and MMDiT~\cite{esser2024scaling}), and $\hat{x}_0 = x_\theta(z,\; t{=}1)$ for an $x_0$-prediction model (\eg, JiT~\cite{li2025back}), assuming $z_{t=1}$ is pure Gaussian noise. A multi-step model used this way initially produces poor samples (\eg, 290 FID), since it was never trained for one-step generation. We simply treat it as a one-step generator and optimize it with the \method. This procedure can transform multi-step generators into competitive one-step generators, without adversarial training or teacher distillation. In this sense, \method provides a minimalistic distribution matching objective.

\subsection{\fdr{k}: Normalized \frechet Distance Ratios under K Models}
\label{sec:fdn}

\paragraph{Metric paradox.}
If measured only by FID~\cite{heusel2017gans}, state-of-the-art image generators appear to have already surpassed real validation images. As shown in Figure~\ref{fig:scatter_benchmark}, when evaluated against ImageNet training images, validation images themselves obtain FID 1.68,\footnote{For ImageNet, FID is computed against \emph{training} set statistics. Therefore, the FID of validation images is not zero.} while many recent works report FID around or below 1.5~\cite{yu2026autoregressive,yu2024representation,zheng2025diffusion,li2025autoregressive,yang2025latent,wang2025ddt,yao2025reconstruction}. Yet, generated images are still clearly distinguishable from real ones. This disconnect between metric and visual quality suggests that FID, which relies on a single, dated feature space, has saturated as a quality signal.

\paragraph{FD ratio.}
A natural remedy is to evaluate across diverse feature spaces.
However, raw FD values are not comparable across representations (\eg, FD-MAE and FD-DINOv2 differ by orders of magnitude; see Tab.~\ref{tab:rawfd_system}). 
To obtain comparable quantities, we normalize the FD of generated images by the FD of \emph{real validation images} in the same feature space. 
Let $\mathcal{T}$ denote the ImageNet training set, $\mathcal{V}$ the validation set, and $\mathcal{G}$ a set of generated images. For a representation model $\phi_i$, we define the \emph{normalized FD ratio}, abbreviated as $\mathrm{FDr}$:
\begin{equation}
  \mathrm{FDr}_{\phi_i}(\mathcal{G})
  =
  \frac{\mathrm{FD}_{\phi_i}(\mathcal{G}, \mathcal{T})}
  {\mathrm{FD}_{\phi_i}(\mathcal{V}, \mathcal{T})}.
  \label{eq:fd_ratio}
  \end{equation}
This ratio is unitless and has a direct interpretation. For example, $\mathrm{FDr}_{\phi_i}=2.0$ means that under $\phi_i$, the generated images are perceptually twice as far from the training set as the validation images. By definition, validation images score exactly $1.0$.

\paragraph{FDr$^K$.} We define \fdr{k} by averaging these normalized ratios over $K$ representation models:
\begin{equation}
    \mathrm{FDr}^{K}(\mathcal{G})
    =
    \frac{1}{K}\sum_{i=1}^{K} \mathrm{FDr}_{\phi_i}(\mathcal{G}).
    \label{eq:fdrk}
\end{equation}
This yields a single multi-representation metric while preserving the interpretation of the per-model ratios. Existing strong generators may beat validation images in Inception~\cite{szegedy2016rethinking} feature space alone, but remain substantially inferior once evaluated across diverse representations (Tabs.~\ref{tab:system},~\ref{tab:detail_system},~\ref{tab:rawfd_system}).

In this paper, we instantiate \fdr{k} with 6 representative models spanning supervised, self-supervised, and vision-language objectives across both CNN and ViT architectures:
  Inception-v3~\cite{szegedy2016rethinking},
  ConvNeXt-v2~\cite{liu2022convnet},
  DINOv2~\cite{oquab2024dinov2},
  MAE~\cite{he2022masked},
  SigLIP2~\cite{tschannen2025siglip}, and
  CLIP~\cite{radford2021learning}.
Details are in Appendix~\ref{sec:appendix_repr_models}.

\paragraph{Scope.}
\fdr{6} should \emph{not} be viewed as a north star, nor as a replacement for human evaluation, but rather as a more robust \emph{automatic} metric than FID alone. It retains the simplicity of \frechet-style evaluation while reducing the blind spots of any single representation. Like any automatic metric, \fdr{k} has its own limitations: its value depends on the choice of $K$ representations, and it still inherits the Gaussian moment-matching assumption of \frechet Distance itself. We view it as a step forward from FID, not a final answer. We provide more discussions in Appendix~\ref{sec:appendix_design_attempts}.

\section{Experiments}
\label{sec:exp}

\paragraph{Setup.}
We study class-conditional image generation on ImageNet-1k~\cite{deng2009imagenet} at $256\times256$ and $512\times 512$ resolutions.
Our experiments cover both pixel-space generators (pMF~\cite{lu2026one}, JiT~\cite{li2025back}) and latent-space generators (iMF~\cite{geng2025improved}). 
All methods are reimplemented and integrated in a unified codebase for fair comparison; all models are initialized from officially released pre-trained weights.
We evaluate using FID~\cite{heusel2017gans} and IS~\cite{Salimans2016} to facilitate comparison with prior work. We also report \fdr{6} as defined in Equation~\ref{eq:fdrk}.
Following standard practice, all metrics are computed from 50,000 generated images against \emph{training} set statistics. We also report metrics for 50,000 \emph{validation} images as a reference.

\paragraph{Training.}
We post-train with a global batch size of 1024 using AdamW~\cite{Loshchilov2019} with a cosine lr schedule and 5 epochs of warm-up. We set $lr{=}10^{-6}$ for pMF~\cite{lu2026one} and iMF~\cite{geng2025improved}, and $lr{=}10^{-5}$ for JiT~\cite{li2025back}. For ablation experiments, we post-train for 50 epochs; for system-level results, 100 epochs.

\subsection{Properties of Population Size in \method}

We first analyze the importance of population size when optimizing \method. Unless stated otherwise, we post-train pMF-B/16~\cite{lu2026one} for 50 epochs using Inception~\cite{szegedy2016rethinking} as the only representation model.

\begin{table*}[t]
\vspace{.1em}
\centering
\subfloat[
\textbf{Queue size}. A large queue is critical for stable FD optimization (\eg, 5--100k), while an overly large queue becomes too stale to remain on-policy.
\label{tab:queue_size}
]{
\begin{minipage}[t]{0.46\linewidth}{\begin{center}
\vspace{-0.5em}
\tablestyle{4pt}{1.05}
\begin{tabular}{y{34}x{24}x{26}y{26}}
queue size & FID$\downarrow$ & IS$\uparrow$ & \fdr{6}$\downarrow$ \\
\shline
\gc{Base} & \gc{3.31} & \gc{254.6} & \gc{13.70} \\
\gc{0k$^\ddagger$} & \gc{3.84} & \gc{250.9} & \gc{17.06} \\
\hline
5k & 1.05 & 280.0 & 11.89 \\
10k & 0.93 & 283.9 & 11.71 \\
\baseline{50k} & \baseline{\textbf{0.89}} & \baseline{288.3} & \baseline{\textbf{10.91}} \\
100k & 0.93 & 288.8 & 11.15 \\
500k & 1.22 & \textbf{294.4} & 17.67 \\
\end{tabular}
\end{center}}\end{minipage}
}
\hspace{1.8em}
\subfloat[
\textbf{EMA decay rate}. The EMA estimator is stable across a wide range ($\beta{=}0.9$ to $0.9999$) and works best at $\beta{=}0.999$.
\label{tab:ema_beta}
]{
\begin{minipage}[t]{0.42\linewidth}{\begin{center}
\vspace{-0.5em}
\tablestyle{4pt}{1.05}
\begin{tabular}{y{26}x{24}x{26}y{26}}
$\beta$ & FID$\downarrow$ & IS$\uparrow$ & \fdr{6}$\downarrow$ \\
\shline
\gc{Base} & \gc{3.31} & \gc{254.6} & \gc{13.70} \\
\gc{0.0$^\ddagger$} & \gc{3.84} & \gc{250.9} & \gc{17.06} \\
\hline
0.9 & 0.98 & 283.6 & 11.19 \\
0.99 & 0.84 & 291.8 & \textbf{10.74} \\
\baseline{0.999} & \baseline{\textbf{0.81}} & \baseline{\textbf{294.5}} & \baseline{10.81} \\
0.9999 & 0.98 & 287.7 & 11.63 \\
\multicolumn{4}{c}{~}\\
\end{tabular}
\end{center}}\end{minipage}
}
\\[0.2em]
\subfloat[
\textbf{Representation model}. Each row optimizes \method in one or more representation spaces; each column evaluates under a different representation. Setting: EMA with $\beta=0.999$.
\label{tab:backbone}
]{
\resizebox{\textwidth}{!}{%
\tablestyle{5pt}{1.05}
\begin{tabular}{l | c c c c c | c c c c}
& \multicolumn{5}{c|}{\fdr (Eq.~\ref{eq:fd_ratio}) $\downarrow$} & \multicolumn{4}{c}{} \\
loss & Incep. & ConvNeXt & DINOv2 & MAE & SigLIP
& FID$\downarrow$ & IS$\uparrow$ & {\small FDr-CLIP}$^\dagger$$\downarrow$ & \fdr{6}$\downarrow$ \\
\shline
\gc{\textit{Validation set images}} & \gc{\textit{1.00}} & \gc{\textit{1.00}} & \gc{\textit{1.00}} & \gc{\textit{1.00}} & \gc{\textit{1.00}} & \gc{\textit{1.68}} & \gc{\textit{232.2}} & \gc{\textit{1.00}} & \gc{\textit{1.00}} \\
\gc{Base} & \gc{1.98} & \gc{1.93} & \gc{10.13} & \gc{13.81} & \gc{31.03} & \gc{3.31} & \gc{254.6} & \gc{23.30} & \gc{13.70} \\
\hline
FD-Inception & \textbf{0.48} & 1.26 & 7.52 & 8.51 & 26.02 & \textbf{0.81} & 294.5 & 21.07 & 10.81 \\
FD-ConvNeXt & 0.98 & \textbf{0.34} & 4.93 & 7.48 & 17.38 & 1.64 & 281.0 & 19.66 & 8.46 \\
FD-DINOv2 & 2.91 & 2.14 & \textbf{2.11} & 10.88 & 16.92 & 4.89 & 347.1 & 15.83 & 8.47 \\
FD-MAE & 3.83 & 1.92 & 5.30 & \textbf{1.11} & 14.44 & 6.42 & 344.0 & 13.19 & 6.63 \\
FD-SigLIP & 4.60 & 2.69 & 4.27 & 9.09 & \textbf{3.61} & 7.71 & \textbf{399.4} & 10.84 & 5.85 \\
\hline
FD-SigLIP+Incep. & 0.53 & 0.95 & 4.99 & 8.66 & 6.83 & 0.89 & 307.5 & 13.75 & 5.95 \\
\baseline{FD-SigLIP+Incep.+MAE (\textbf{SIM})} & \baseline{0.56} & \baseline{0.85} & \baseline{4.65} & \baseline{2.36} & \baseline{6.94} & \baseline{0.94} & \baseline{307.8} & \baseline{\textbf{9.81}} & \baseline{\textbf{4.20}} \\
\end{tabular}%
}
}
\vspace{-.5em}
\caption{
\textbf{Properties of \method.}
Ablation results on pMF-B/16~\cite{lu2026one} post-trained for 50 epochs.
(a,\,b) study population size; (c) studies representation model choice.
Base: base model \emph{before} post-training.
$^\ddagger$Statistics estimated from the current batch only (batch size 1024).
$^\dagger$FDr-CLIP is reported separately because CLIP is never used as a training signal in any row shown here; it is included in \fdr{6}.
Default setting is marked in \baselinelegend{Gray}.
}
\vspace{-1.1em}
\end{table*}

\paragraph{Population size via queue size.} Table~\ref{tab:queue_size} studies the effect of the queue size $N$.
Without a queue ($N{=}0$), statistics are estimated from the current batch only, which degrades all metrics relative to the base model (FID: 3.31$\to$3.84, \fdr{6}: 13.70$\to$17.06).
Performance improves steadily when using queue size from 5k to 50k (FID 0.89, \fdr{6} 10.91), but degrades beyond 100k as cached features become increasingly off-policy, and the stale statistics outweigh the benefit of a larger population.
Notably, at 500k the queue becomes overly stale that FID and \fdr{6} \emph{disagree}: FID still improves over the base model (1.22 \vs 3.31), whereas \fdr{6} degrades beyond the base model (17.67 \vs 13.70). This is an early sign that FID alone can be misleading.

\paragraph{Population size via EMA decay rate.}
Table~\ref{tab:ema_beta} studies the EMA estimator, where the decay rate $\beta$ implicitly controls the effective population size.
The estimator is robust across a wide range ($\beta{=}0.9$ to $0.9999$). The best setting, $\beta{=}0.999$, achieves 0.81 FID and 10.81 \fdr{6}, improving the best queue result while requiring negligible extra memory.
We use $\beta{=}0.999$ in all subsequent experiments.

Both studies confirm that \method needs a population larger than the optimization batch, but not so large that staleness dominates. We default to EMA for its simplicity and stronger results.

\subsection{Properties of Representation Models in \method}

\newcommand{\reprida}{000015}
\newcommand{\repridb}{015289}
\newcommand{\repridc}{000098}
\newcommand{\repridd}{038232} %
\newcommand{\repride}{000331}
\newcommand{\repridg}{015207}
\newcommand{\repridh}{004466}
\newcommand{\repridi}{005000}
\newcommand{\reprrowa}{\reprida}
\newcommand{\reprrowb}{\repridb}
\newcommand{\reprrowc}{\repridc}
\newcommand{\reprrowd}{\repridd}
\newcommand{\reprrowe}{\repride}
\newcommand{\reprrowf}{\repridf}
\newcommand{\reprrowg}{\repridg}
\newcommand{\reprrowh}{\repridh}
\newcommand{\reprrowi}{\repridi}
\newcommand{\reprrowj}{\repridj}
\definecolor{fidcolor}{RGB}{52,168,83}     %
\providecommand{\ttiny}{\fontsize{4pt}{5pt}\selectfont}
\providecommand{\xsmall}{\fontsize{8pt}{9pt}\selectfont}
\definecolor{fdrcolor}{RGB}{251,188,4}     %
\newcommand{\metricval}[2]{%
  \begingroup\setlength{\fboxsep}{1pt}%
  \colorbox{#1}{\raisebox{0pt}[1.1ex][.15ex]{#2}}%
  \endgroup}
\newcommand{\reprcolheader}[5]{%
  \begin{tabular}[b]{@{}c@{}}%
    \xsmall #1\\[0.1em]%
    \tiny\begin{tabular}{@{}r@{\,}l@{}}%
      FID & \metricval{fidcolor!#4}{#2}\\[0.15em]%
      \fdr{6} & \metricval{fdrcolor!#5}{#3}%
    \end{tabular}%
  \end{tabular}%
}
\newcommand{\reprcola}{\reprcolheader{Original}{3.31}{13.70}{38}{8}}
\newcommand{\reprdira}{base}

\newcommand{\reprcolb}{\reprcolheader{Inception}{0.81}{10.81}{80}{13}}
\newcommand{\reprdirb}{fd_inception}

\newcommand{\reprcolc}{\reprcolheader{ConvNeXt}{1.64}{8.46}{55}{28}}
\newcommand{\reprdirc}{fd_convnext}
\newcommand{\reprcold}{\reprcolheader{DINOv2}{4.89}{8.47}{28}{24}}
\newcommand{\reprdird}{fd_dinov2}
\newcommand{\reprcole}{\reprcolheader{MAE}{6.42}{6.63}{18}{40}}
\newcommand{\reprdire}{fd_mae}
\newcommand{\reprcolf}{\reprcolheader{SigLIP}{7.71}{5.85}{8}{50}}
\newcommand{\reprdirf}{fd_siglip}
\newcommand{\reprcolh}{\reprcolheader{\tiny SigLIP+Incep.+MAE}{0.94}{4.20}{68}{60}}
\newcommand{\reprdirh}{fd_siglip_mae_inception}
\newcommand{\reprvispad}{0pt}        %
\newcommand{\reprvisrowsep}{0pt}     %
\newcommand{\reprbandgap}{3pt}       %
\newcommand{\reprviswidth}{0.13\textwidth} %

\begin{figure*}[t]
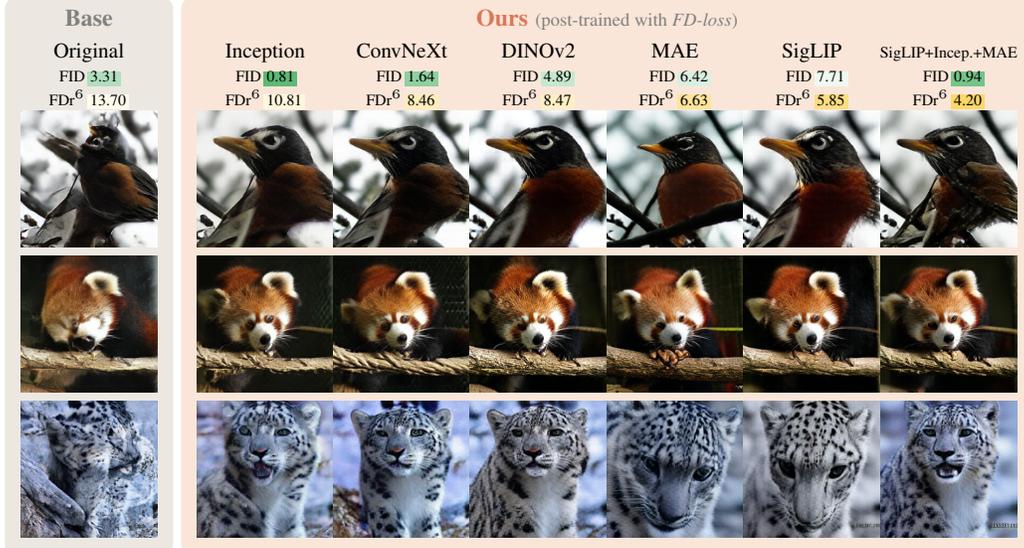

\begin{center}
{\newcommand{\reprvisimg}[2]{\includegraphics[width=\reprviswidth]{figures/repr_single/#1/#2.jpg}}%
\setlength{\tabcolsep}{0pt}%
\renewcommand{\arraystretch}{1}%
\newcommand{\basecolblock}{%
  \begin{tabular}{@{}c@{}}
    \textbf{\small\textcolor{gray}{Base}}\\[1pt]
    \reprcola\\[1pt]
    \reprvisimg{\reprdira}{\reprrowa}\\[\reprvisrowsep]
    \reprvisimg{\reprdira}{\reprrowc}\\[\reprvisrowsep]
    \reprvisimg{\reprdira}{\reprrowe}\\
  \end{tabular}%
}
\newcommand{\ourscolblock}{%
  \begin{tabular}{@{}c@{\hspace{\reprvispad}}c@{\hspace{\reprvispad}}c@{\hspace{\reprvispad}}c@{\hspace{\reprvispad}}c@{\hspace{\reprvispad}}c@{}}
    \multicolumn{6}{c}{\textbf{\small\textcolor{accentorange}{Ours}}\;{\scriptsize\textcolor{gray}{(post-trained with \method)}}}\\[1pt]
    \reprcolb & \reprcolc & \reprcold & \reprcole & \reprcolf & \reprcolh\\[1pt]
    \reprvisimg{\reprdirb}{\reprrowa} & \reprvisimg{\reprdirc}{\reprrowa} & \reprvisimg{\reprdird}{\reprrowa} & \reprvisimg{\reprdire}{\reprrowa} & \reprvisimg{\reprdirf}{\reprrowa} & \reprvisimg{\reprdirh}{\reprrowa}\\[\reprvisrowsep]
    \reprvisimg{\reprdirb}{\reprrowc} & \reprvisimg{\reprdirc}{\reprrowc} & \reprvisimg{\reprdird}{\reprrowc} & \reprvisimg{\reprdire}{\reprrowc} & \reprvisimg{\reprdirf}{\reprrowc} & \reprvisimg{\reprdirh}{\reprrowc}\\[\reprvisrowsep]
    \reprvisimg{\reprdirb}{\reprrowe} & \reprvisimg{\reprdirc}{\reprrowe} & \reprvisimg{\reprdird}{\reprrowe} & \reprvisimg{\reprdire}{\reprrowe} & \reprvisimg{\reprdirf}{\reprrowe} & \reprvisimg{\reprdirh}{\reprrowe}\\
  \end{tabular}%
}
\makebox[\textwidth][c]{%
  \tikz[baseline]{\node[fill=basebg, rounded corners=4pt, inner xsep=6pt, inner ysep=3pt, line width=0pt] {\basecolblock};}%
  \hspace{\reprbandgap}%
  \tikz[baseline]{\node[fill=oursbg, rounded corners=4pt, inner xsep=6pt, inner ysep=3pt, line width=0pt] {\ourscolblock};}%
}
}%
\vspace{-.25em}
\caption{
\textbf{\method improves visual quality under different representations.}
Samples from pMF-B/16~\cite{lu2026one} post-trained with \method.
Darker {\setlength{\fboxsep}{1pt}\colorbox{fidcolor!60}{\raisebox{0pt}[1.1ex][.15ex]{green}}}: lower FID; darker {\setlength{\fboxsep}{1pt}\colorbox{fdrcolor!60}{\raisebox{0pt}[1.1ex][.15ex]{yellow}}}: lower \fdr{6}.
Post-trained models improve over the base model (left). Inception post-trained model achieves the lowest FID (0.81) yet does not produce the best samples; models post-trained with modern representations achieve lower \fdr{6} and show better object structure despite higher FID.
}
\label{fig:repr_single}
\vspace{-1em}
\end{center}
\end{figure*}

\paragraph{Single representation model.}%
Table~\ref{tab:backbone} studies \method under different representations; Figure~\ref{fig:repr_single} shows qualitative comparisons. 
As expected, each model scores best in the representation space it optimizes (on-diagonal).
However, the off-diagonal behavior differs across model families. Optimizing Inception~\cite{szegedy2016rethinking} gives the best FID (3.31$\to$0.81) and improves \fdr{6} slightly (13.70$\to$10.81). Optimizing modern ViTs, \eg, DINOv2, MAE, SigLIP2, \emph{worsens} FID but \emph{improves} \fdr{6} more substantially. ConvNeXt~\cite{liu2022convnet}, a modern CNN, sits in between: it improves FID less than Inception (3.31$\to$1.64) but \fdr{6} more (13.70$\to$8.46). This reveals that CNN-based representations tend to improve FID, while ViT-based ones improve the broader \fdr{6} more.

These results already indicate that lower FID and better overall quality are \emph{not} always the same objective. Improvements captured by modern feature spaces can be invisible, or even unfavorable, under Inception. Moreover, under Inception~\cite{szegedy2016rethinking} and ConvNeXt~\cite{liu2022convnet}, \method can even drive \fdr{} below 1.0; in other words, the generated images become statistically closer to the training set than real validation images, suggesting that some feature spaces are easier to saturate than others.

Figure~\ref{fig:repr_single} further supports this. Post-trained models improve visually over the base model. 
Yet, the lowest-FID model (Inception, 0.81) does not produce the best samples; instead, models post-trained with modern representations show better object structure despite much higher FID.

\paragraph{Multiple representation models.}
Table~\ref{tab:backbone} further studies combinations of representations (Eq.~\ref{eq:multi}).
Combining representations is generally more effective than using a single one.
Optimizing SigLIP with Inception recovers FID to 0.89 while maintaining strong \fdr{6}.
Adding MAE (denoted FD-SIM) further improves \fdr{6} with a negligible FID trade-off.
We use FD-SIM as the default from now on.

\subsection{Repurposing Multi-Step Generators into One-Step Generators}

\newcommand{\repurposeimga}{00016_rank02_class0207}
\newcommand{\repurposeimgb}{00100_rank12_class0088}
\newcommand{\repurposeimgc}{00132_rank16_class0088}
\newcommand{\repurposeimgd}{00466_rank58_class0387}
\newcommand{\repurposeimge}{00447_rank55_class0279}
\newcommand{\repurposeimgf}{00047_rank05_class0279}
\newcommand{\repurposeimgg}{00431_rank53_class0279}
\newcommand{\repurposeimgh}{00018_rank02_class0387}
\newcommand{\repurposeimgi}{00024_rank03_class0207}
\newcommand{\repurposeimgj}{00080_rank10_class0207}

\begin{figure*}[t]
\vspace{-0.5em}
\begin{center}
\newcommand{\repimgw}{0.13\textwidth}
\newcommand{\reppad}{4pt}
\newcommand{\repimg}[2]{\includegraphics[width=\repimgw]{figures/repurpose_jit/#1/#2.jpg}}
\newcommand{\repsep}{\hspace{3pt}\textcolor{gray!40}{\vrule width 0.4pt}\hspace{0pt}}
\newcommand{\rephead}[1]{\makebox[\repimgw][c]{#1}}
\newcommand{\reprow}[1]{%
  \makebox[\linewidth][c]{%
    \hspace{\reppad}\repimg{base_1step}{#1}%
    \hspace{\reppad}\repimg{base_50step}{#1}%
    \repsep%
    \hspace{\reppad}\repimg{fd_inception}{#1}%
    \hspace{\reppad}\repimg{fd_mae}{#1}%
    \hspace{\reppad}\repimg{fd_siglip_mae}{#1}%
    \hspace{\reppad}\repimg{fd_siglip_mae_inception}{#1}%
  }\par\vspace{0.1pt}%
}
\begin{minipage}[c]{0.45\textwidth}
\centering
\captionof{table}{
\textbf{\method repurposes multi-step JiT models to generate in one step.}
All post-trained models use 1 NFE. Setting: JiT-L/16~\cite{li2025back}, post-trained for 50 epochs.
$^\dagger$200 NFE $=$ 50 steps $\times$ 2 (Heun) $\times$ 2 (CFG).
}
\label{tab:repurpose}
\vspace{.3em}
\resizebox{\linewidth}{!}{%
\tablestyle{4pt}{1.05}
\begin{tabular}{l c c c c}
setting & NFE & FID$\downarrow$ & IS$\uparrow$ & \fdr{6}$\downarrow$ \\
\shline
\multicolumn{5}{l}{\hspace{-.5em} \textit{base model}} \\
\gc{JiT-L (50-step)} & \gc{200$^\dagger$} & \gc{2.59} & \gc{288.5} & \gc{10.73} \\
\gc{JiT-L (1-step)} & \gc{1} & \gc{291.59} & \gc{2.0} & \gc{214.75} \\
\hline
\multicolumn{5}{l}{\hspace{-.5em} \textit{models post-trained with \method}} \\
FD-Incep. & 1 & \textbf{0.77} & 293.7 & 12.86 \\
FD-MAE & 1 & 6.52 & 280.4 & 9.30 \\
FD-SigLIP & 1 & 5.10 & 329.6 & 9.04 \\
FD-SigLIP+MAE & 1 & 4.67 & \textbf{354.0} & 3.83 \\
\baseline{FD-SigLIP+Incep.+MAE (\textbf{SIM})} & \baseline{1} & \baseline{0.85} & \baseline{319.5} & \baseline{\textbf{3.29}} \\
\end{tabular}}
\end{minipage}%
\hfill
\begin{minipage}[c]{0.51\textwidth}
\centering
\newcommand{\repimgcol}[1]{\includegraphics[width=\repimgw]{figures/repurpose_jit/#1}}
{\setlength{\tabcolsep}{0pt}%
\newcommand{\repbasetab}{%
  \begin{tabular}{@{}cc@{}}
    \multicolumn{2}{c}{\textbf{\scriptsize\textcolor{gray}{Base}}} \\[-1pt]
    {\tiny 1-step} & {\tiny 50-step} \\[0pt]
    \repimgcol{base_1step/\repurposeimga.jpg} & \repimgcol{base_50step/\repurposeimga.jpg} \\[-2pt]
    \repimgcol{base_1step/\repurposeimgc.jpg} & \repimgcol{base_50step/\repurposeimgc.jpg} \\[-2pt]
    \repimgcol{base_1step/\repurposeimgd.jpg} & \repimgcol{base_50step/\repurposeimgd.jpg} \\[-2pt]
    \repimgcol{base_1step/\repurposeimgg.jpg} & \repimgcol{base_50step/\repurposeimgg.jpg} \\
  \end{tabular}%
}
\newcommand{\repourstab}{%
  \begin{tabular}{@{}cccc@{}}
    \multicolumn{4}{c}{\textbf{\scriptsize\textcolor{accentorange}{Ours}}\;{\tiny\textcolor{gray}{(post-trained with \method)}}} \\[-1pt]
    {\tiny Inception} & {\tiny MAE} & {\fontsize{4pt}{5pt}\selectfont SigLIP+MAE} & {\fontsize{3pt}{4pt}\selectfont SigLIP+Incep.+MAE} \\[0pt]
    \repimgcol{fd_inception/\repurposeimga.jpg} & \repimgcol{fd_mae/\repurposeimga.jpg} & \repimgcol{fd_siglip_mae/\repurposeimga.jpg} & \repimgcol{fd_siglip_mae_inception/\repurposeimga.jpg} \\[-2pt]
    \repimgcol{fd_inception/\repurposeimgc.jpg} & \repimgcol{fd_mae/\repurposeimgc.jpg} & \repimgcol{fd_siglip_mae/\repurposeimgc.jpg} & \repimgcol{fd_siglip_mae_inception/\repurposeimgc.jpg} \\[-2pt]
    \repimgcol{fd_inception/\repurposeimgd.jpg} & \repimgcol{fd_mae/\repurposeimgd.jpg} & \repimgcol{fd_siglip_mae/\repurposeimgd.jpg} & \repimgcol{fd_siglip_mae_inception/\repurposeimgd.jpg} \\[-2pt]
    \repimgcol{fd_inception/\repurposeimgg.jpg} & \repimgcol{fd_mae/\repurposeimgg.jpg} & \repimgcol{fd_siglip_mae/\repurposeimgg.jpg} & \repimgcol{fd_siglip_mae_inception/\repurposeimgg.jpg} \\
  \end{tabular}%
}
\resizebox{\linewidth}{!}{%
  \tikz[baseline]{\node[fill=basebg, rounded corners=3pt, inner xsep=5pt, inner ysep=2pt, line width=0pt] {\repbasetab};}%
  \hspace{3pt}%
  \tikz[baseline]{\node[fill=oursbg, rounded corners=3pt, inner xsep=5pt, inner ysep=2pt, line width=0pt] {\repourstab};}%
}}
\end{minipage}
\vspace{-0.6em}
\caption{
\textbf{Repurposing a multi-step model into a one-step generator with \method.}
Samples from the \emph{same} noise input across the base model and different post-trained models. The naive one-step base model fails to produce sensible images.
After post-training, the 1-NFE models generate sensible images, and the strongest variants are visually comparable or superior to the 50-step base model.
}
\label{fig:repurpose_jit}
\vspace{-1.7em}
\end{center}
\end{figure*}

We study whether \method can repurpose a pre-trained multi-step model into a one-step generator.
We use JiT-L/16~\cite{li2025back} as the base model and post-train it for 50 epochs following Section~\ref{sec:training_setup}.
Table~\ref{tab:repurpose} and Figure~\ref{fig:repurpose_jit} report the results.
As expected, the base model fails in the naive one-step setting, since it is trained for multi-step denoising.
After post-training with \method, all variants produce sensible images.
FD-SIM achieves the best \fdr{6} (3.29) with FID 0.85, while FD-Inception gives the best FID (0.77) but a much higher \fdr{6} (12.86).
Notably, repurposing is more demanding than improving an already strong one-step model: capable representations such as MAE, SigLIP2, and their combinations all yield visually compelling samples, whereas Inception alone might not be strong enough, so the repurposed models may exhibit certain artifacts.
These results indicate that the same \method recipe works both for improving one-step generators and for converting multi-step ones into one-step, with no distillation, adversarial loss, or per-sample regression target.

\subsection{Comparisons}

\begin{table}[t]
\caption{
\textbf{\method generalizes across model families, sizes, and image resolutions.}
(a)--(c) cover three generator families on ImageNet at 256px; (d) covers pMF on ImageNet 512px. Each sub-table shows the base generator and its \method post-trained variants (\baselinelegend{shaded}). All post-trained models use 1 NFE. SIM: SigLIP+Inception+MAE. Setting: post-trained for 100 epochs, EMA with $\beta=0.999$.
}
\label{tab:all_fd_models}
\vspace{-5pt}
\centering
\scriptsize
\captionsetup[subtable]{font=scriptsize,skip=2pt}
\begin{subtable}[t]{0.25\linewidth}
\centering
\tablestyle{1.5pt}{1.03}
\caption{pMF~\cite{lu2026one} (pixel, 256px)}\label{tab:all_fd_pmf256}
\begin{tabular}{@{}l c c@{}}
method & FID & \fdr{6} \\
\shline
pMF-B & 3.31 & 13.70 \\
\rowcolor{baselinecolor} \ +\,Incep. & \textbf{0.77} & 10.66 \\
\rowcolor{baselinecolor} \ +\,SIM & 0.85 & \textbf{3.50} \\
\hline
pMF-L & 2.72 & 9.09 \\
\rowcolor{baselinecolor} \ +\,Incep. & \textbf{0.73} & 6.19 \\
\rowcolor{baselinecolor} \ +\,SIM & 0.78 & \textbf{2.09} \\
\hline
pMF-H & 2.29 & 6.87 \\
\rowcolor{baselinecolor} \ +\,Incep. & \textbf{0.72} & 4.86 \\
\rowcolor{baselinecolor} \ +\,SIM & 0.77 & \textbf{1.89} \\
\end{tabular}
\end{subtable}
\hfill
\begin{subtable}[t]{0.25\linewidth}
\centering
\tablestyle{1.5pt}{1.03}
\caption{iMF~\cite{geng2025improved} (latent, 256px)}\label{tab:all_fd_imf256}
\begin{tabular}{@{}l c c@{}}
method & FID & \fdr{6} \\
\shline
iMF-B & 3.45 & 15.29 \\
\rowcolor{baselinecolor} \ +\,Incep. & \textbf{0.79} & 11.34 \\
\rowcolor{baselinecolor} \ +\,SIM & 0.88 & \textbf{5.56} \\
\hline
iMF-L & 1.93 & 9.06 \\
\rowcolor{baselinecolor} \ +\,Incep. & \textbf{0.75} & 6.63 \\
\rowcolor{baselinecolor} \ +\,SIM & 0.79 & \textbf{2.74} \\
\hline
iMF-XL & 1.82 & 8.39 \\
\rowcolor{baselinecolor} \ +\,Incep. & \textbf{0.72} & 6.01 \\
\rowcolor{baselinecolor} \ +\,SIM & 0.76 & \textbf{2.45} \\
\end{tabular}
\end{subtable}
\hfill
\begin{subtable}[t]{0.25\linewidth}
\centering
\tablestyle{1.5pt}{1.03}
\caption{JiT~\cite{li2025back} (pixel, 256px)}\label{tab:all_fd_jit256}
\begin{tabular}{@{}l c c@{}}
method & FID & \fdr{6} \\
\shline
JiT-B & 3.71 & 15.65 \\
\rowcolor{baselinecolor} \ +\,Incep. & \textbf{0.76} & 22.48 \\
\rowcolor{baselinecolor} \ +\,SIM & 1.00 & \textbf{5.53} \\
\hline
JiT-L & 2.59 & 10.73 \\
\rowcolor{baselinecolor} \ +\,Incep. & \textbf{0.73} & 12.75 \\
\rowcolor{baselinecolor} \ +\,SIM & 0.77 & \textbf{3.24} \\
\hline
JiT-H & 1.97 & 7.66 \\
\rowcolor{baselinecolor} \ +\,Incep. & \textbf{0.72} & 10.18 \\
\rowcolor{baselinecolor} \ +\,SIM & 0.75 & \textbf{2.65} \\
\end{tabular}
\end{subtable}
\hfill
\begin{subtable}[t]{0.2\linewidth}
\centering
\tablestyle{1.5pt}{1.03}
\caption{pMF~\cite{lu2026one} (pixel, 512px)}\label{tab:all_fd_pmf512}
\begin{tabular}{@{}l c c@{}}
method & FID & \fdr{6} \\
\shline
pMF-B & 3.59 & 15.04 \\
\rowcolor{baselinecolor} \ +\,SIM & \textbf{0.87} & \textbf{3.82} \\
\hline
pMF-L & 2.56 & 9.76 \\
\rowcolor{baselinecolor} \ +\,SIM & \textbf{0.80} & \textbf{2.12} \\
\hline
pMF-H & 2.43 & 7.33 \\
\rowcolor{baselinecolor} \ +\,SIM & \textbf{0.78} & \textbf{1.81} \\
\end{tabular}
\end{subtable}
\vspace{-2em}
\end{table}

\begin{table*}[t]
\begin{center}{
\caption{
\textbf{System-level comparison on ImageNet $256\times256$.}
All metrics for all methods are computed by us under a unified evaluation pipeline; numbers may differ slightly from the original papers.
Our \method (\baselinelegend{shaded} rows) improves already strong generators across geneartor families and model scales. 
$^\dagger$CFG is applied only in a time sub-interval; we report the full-CFG upper bound for simplicity.
Uncurated qualitative samples are in Appendix~\ref{sec:appendix_samples}.
}
\label{tab:system}
\vspace{-.2em}
\tablestyle{4pt}{1.03}
\begin{tabular}{l l c c c c c c c}
method & NFE & space & \#params & \fdr{6}$\downarrow$ & FID$\downarrow$ & IS$\uparrow$ & Prec$\uparrow$ & Recall$\uparrow$ \\
\shline
\multicolumn{9}{l}{\hspace{-.5em} \textit{reference (real images)}} \\
\gc{50k validation images} & \gc{N/A} & \gc{N/A} & \gc{N/A} & \gc{1.00} & \gc{1.68} & \gc{232.2} & \gc{0.75} & \gc{0.66} \\
\hline
\multicolumn{9}{l}{\hspace{-.5em} \textit{discrete-space models}} \\
VAR-d30 & 10$\times$2 & discrete & 2B & 6.70 & 1.97 & \textbf{304.6} & 0.82 & 0.59 \\
BAR-L~\cite{yu2026autoregressive} & 256$\times$2{\tiny$\times$4} & discrete & 1.1B & \textbf{3.57} & \textbf{1.01} & 281.9 & 0.77 & 0.68 \\
\hline
\multicolumn{9}{l}{\hspace{-.5em} \textit{latent-space models, multi-step}} \\
\multicolumn{9}{l}{\hspace{-.2em} \textit{without semantic distillation}} \\
SiT-XL/2~\cite{ma2024sit} & 250$\times$2 & latent & 675M & 8.44 & 2.12 & 256.7 & 0.81 & 0.60 \\
MAR-L~\cite{li2025autoregressive} & 256$\times$2{\tiny$\times$100} & latent & 478M & 6.68 & 1.80 & 293.4 & 0.80 & 0.60 \\
FlowAR-H~\cite{ren2024flowar} & 50$\times$2$^\dagger$ & latent & 1.9B & 6.13 & 1.68 & 274.1 & 0.80 & 0.62 \\
MAR-H~\cite{li2025autoregressive} & 256$\times$2{\tiny $\times$100} & latent & 942M & 5.61 & 1.56 & 299.5 & 0.80 & 0.62 \\
MAR-L, DeTok~\cite{yang2025latent} & 256$\times$2{\tiny $\times$100} & latent & 478M & \textbf{5.49} & \textbf{1.39} & \textbf{306.2} & 0.81 & 0.62 \\
\multicolumn{9}{l}{\hspace{-.2em} \textit{with semantic distillation}} \\
REG~\cite{wu2025representation} & 250$\times$2$^\dagger$ & latent & 685M & 4.64 & 1.54 & 302.9 & 0.78 & 0.62 \\
SiT-XL/2-REPA~\cite{yu2024representation} & 250$\times$2$^\dagger$ & latent & 675M & 5.45 & 1.42 & 306.1 & 0.80 & 0.65 \\
LightningDiT~\cite{yao2025reconstruction} & 250$\times$2 & latent & 675M & 4.57 & 1.42 & 294.3 & 0.80 & 0.64 \\
DDT-XL~\cite{wang2025ddt} & 250$\times$2 & latent & 675M & 5.70 & 1.26 & \textbf{309.3} & 0.79 & 0.66 \\
REPA-E~\cite{leng2025repa} & 250$\times$2$^\dagger$ & latent & 676M & \textbf{3.04} & 1.17 & 298.3 & 0.79 & 0.66 \\
RAE-XL~\cite{zheng2025diffusion} & 50$\times$2$^\dagger$ & latent & 839M & 3.26 & \textbf{1.16} & 261.0 & 0.77 & 0.67 \\

\hline
\multicolumn{9}{l}{\hspace{-.5em} \textit{latent-space models, one-step}} \\
Drift-L (latent)~\cite{deng2026generative} & 1 & latent & 463M & 10.92 & 1.53 & 257.2 & 0.79 & 0.63 \\
iMF-XL \cite{geng2025improved} & 1 & latent & 610M & 8.39 & 1.82 & 278.9 & 0.78 & 0.63 \\
iMF-XL \cite{geng2025improved} & 2 & latent & 610M & 7.48 & 1.61 & 289.1 & 0.79 & 0.63 \\
\rowcolor{baselinecolor} \quad + \method & 1 & latent & 610M & \textbf{2.45} & \textbf{0.76} & 301.3 & 0.77 & 0.67 \\
\hline
\multicolumn{9}{l}{\hspace{-.5em} \textit{pixel-space models, multi-step}} \\
PixNerd-XL~\cite{wang2026pixnerd} & 100$\times$2 & pixel & 1.0B & 5.01 & 2.10 & \textbf{318.8} & 0.81 & 0.59 \\
JiT-L \cite{li2025back} & 50$\times$2$\times$2$^\dagger$ & pixel & 459M & 10.73 & 2.59 & 288.5 & 0.79 & 0.59 \\
\rowcolor{baselinecolor} \quad + \method & 1 & pixel & 459M & 3.24 & 0.77 & 317.3 & 0.77 & 0.66 \\
JiT-H \cite{li2025back} & 50$\times$2$\times$2$^\dagger$ & pixel & 953M & 7.66 & 1.97 & 296.0 & 0.78 & 0.63 \\
\rowcolor{baselinecolor} \quad + \method & 1 & pixel & 953M & \textbf{2.65} & \textbf{0.75} & 313.0 & 0.76 & 0.66 \\
\hline
\multicolumn{9}{l}{\hspace{-.5em} \textit{pixel-space models, one-step}} \\
Drift-L (pixel)~\cite{deng2026generative} & 1 & pixel & 465M & 10.51 & 1.43 & 305.8 & 0.81 & 0.60 \\
pMF-L \cite{lu2026one} & 1 & pixel & 410M & 9.09 & 2.72 & 261.7 & 0.81 & 0.56 \\
\rowcolor{baselinecolor} \quad + \method & 1 & pixel & 410M & 2.09 & 0.78 & 309.2 & 0.76 & 0.67 \\
pMF-H \cite{lu2026one} & 1 & pixel & 935M & 6.87 & 2.29 & 267.2 & 0.80 & 0.59 \\
\rowcolor{baselinecolor} \quad + \method & 1 & pixel & 935M & \textbf{1.89} & 0.77 & \textbf{310.1} & 0.77 & 0.68 \\
\end{tabular}
}
\end{center}
\vspace{-1em}
\end{table*}

\paragraph{Scalability.}
Table~\ref{tab:all_fd_models} reports \method post-trained models on three generator families (pMF~\cite{lu2026one}, iMF~\cite{geng2025improved}, JiT~\cite{li2025back}), each at three model sizes, on ImageNet $256{\times}256$, and on pMF at three sizes on ImageNet $512{\times}512$. Across all configurations, FD-SIM drives \fdr{6} to between 1.81 and 5.56, and FD-Inception drives FID to between 0.72 and 0.79. These gains are obtained with the same hyperparameters. Per-model tuning could yield better results.
\method thus transfers across pixel and latent spaces, one-step and repurposed multi-step generators, model scales, and image resolutions.

\paragraph{System-level comparison.}
As a reference, we compare \method post-trained models with prior work in Table~\ref{tab:system}.
Since \fdr{6} has not been reported before, we re-sample 50k images from official checkpoints using official code, and re-evaluate \emph{all} methods under the same pipeline.
Our \method post-trained models push FID below all prior systems and drive \fdr{6} substantially lower. 
More importantly, they achieve this with only a \emph{single} network function evaluation (1 NFE).

\paragraph{Human preference.}
Since automatic metrics are only proxies for perceptual quality, we further conduct a pairwise human preference study (Fig.~\ref{fig:human_pref}; Appendix~\ref{sec:appendix_human_pref}).
The study provides two signals.
First, \method post-trained models are preferred over their corresponding base models across all three generator families.
Second, even the strongest generator tested in this study is still preferred less often than real validation images, corroborating that ImageNet generation is not yet solved (Fig.~\ref{fig:scatter_benchmark}).
\providecommand{\ttiny}{\fontsize{4pt}{5pt}\selectfont}
\newcommand{\HPairAW}{75.7}\newcommand{\HPairAL}{24.3}%
\newcommand{\HPairBW}{77.05}\newcommand{\HPairBL}{22.95}%
\newcommand{\HPairCW}{62.3}\newcommand{\HPairCL}{37.7}%
\newcommand{\RealRAEW}{30.1}\newcommand{\RealRAEL}{69.9}%
\newcommand{\RealBARW}{26.4}\newcommand{\RealBARL}{73.6}%
\newcommand{\RealOursW}{37.4}\newcommand{\RealOursL}{62.6}%
\newcommand{\HPBarText}[3]{%
  \node[font=\tiny, inner sep=0pt, text=#3] at (#1)
    {\pgfmathprintnumber[fixed,zerofill,precision=1]{#2}\%};%
}
\pgfmathsetmacro{\LAOurs}{\HPairAW}\pgfmathsetmacro{\LAOpp}{\HPairAL}
\pgfmathsetmacro{\LBOurs}{\HPairBW}\pgfmathsetmacro{\LBOpp}{\HPairBL}
\pgfmathsetmacro{\LCOurs}{\HPairCW}\pgfmathsetmacro{\LCOpp}{\HPairCL}
\pgfmathsetmacro{\RAEGen}{\RealRAEW}\pgfmathsetmacro{\RAEVal}{\RealRAEL}
\pgfmathsetmacro{\BARGen}{\RealBARW}\pgfmathsetmacro{\BARVal}{\RealBARL}
\pgfmathsetmacro{\OursGen}{\RealOursW}\pgfmathsetmacro{\OursVal}{\RealOursL}

\begin{figure}[h!]
\centering
\vspace{-5pt}
\begin{minipage}[t]{0.5\textwidth}
\centering
\begin{tikzpicture}
\begin{axis}[
  xbar stacked,
  set layers=axis on top,
  clip=false,
  width=0.9\textwidth,
  height=3.6cm,
  bar width=12pt,
  y dir=reverse,
  ymin=0.4, ymax=3.6,
  enlargelimits=false,
  xmin=0, xmax=100,
  xtick={0,50,100},
  xticklabels={,,},
  ytick=\empty,
  title={Ours (left) vs. Base Models (right)},
  title style={font=\scriptsize, yshift=-6pt},
  axis line style={axisgray!60, line width=0.4pt},
  tick style={axisgray!60, line width=0.3pt},
]
\draw[reflinecolor, densely dashed, line width=0.6pt]
  (rel axis cs:0.5,0) -- (rel axis cs:0.5,1);

\addplot+[fill=accentorange!85, draw=accentorange!60!black, line width=0.3pt]
coordinates { (\LCOurs,1) (\LBOurs,2) (\LAOurs,3) };
\addplot+[fill=slateblue!75, draw=slateblue!60!black, line width=0.3pt]
coordinates { (\LCOpp,1) (\LBOpp,2) (\LAOpp,3) };

\node[font=\tiny\bfseries, anchor=east, text=accentorange!70!black] at (axis cs:1, 1) {Ours, {\ttiny JiT-H$^*$}};
\node[font=\tiny\bfseries, anchor=east, text=accentorange!70!black] at (axis cs:1, 2) {Ours, {\ttiny iMF-XL$^*$}};
\node[font=\tiny\bfseries, anchor=east, text=accentorange!70!black] at (axis cs:1, 3) {Ours, {\ttiny pMF-H$^*$}};

\node[font=\tiny, anchor=west] at (axis cs:101, 1) {JiT-H$^\dagger$};
\node[font=\tiny, anchor=west] at (axis cs:101, 2) {iMF-XL};
\node[font=\tiny, anchor=west] at (axis cs:101, 3) {pMF-H};

\coordinate (LCw) at (axis cs:{\LCOurs/2},1);
\coordinate (LCl) at (axis cs:{\LCOurs + \LCOpp/2},1);
\coordinate (LBw) at (axis cs:{\LBOurs/2},2);
\coordinate (LBl) at (axis cs:{\LBOurs + \LBOpp/2},2);
\coordinate (LAw) at (axis cs:{\LAOurs/2},3);
\coordinate (LAl) at (axis cs:{\LAOurs + \LAOpp/2},3);
\end{axis}
\HPBarText{LAw}{\LAOurs}{white}
\HPBarText{LAl}{\LAOpp}{white}
\HPBarText{LBw}{\LBOurs}{white}
\HPBarText{LBl}{\LBOpp}{white}
\HPBarText{LCw}{\LCOurs}{white}
\HPBarText{LCl}{\LCOpp}{white}
\end{tikzpicture}
\end{minipage}%
\hfill
\begin{minipage}[t]{0.5\textwidth}
\centering
\begin{tikzpicture}
\begin{axis}[
  xbar stacked,
  set layers=axis on top,
  clip=false,
  width=0.9\textwidth,
  height=3.6cm,
  bar width=12pt,
  y dir=reverse,
  ymin=0.4, ymax=3.6,
  enlargelimits=false,
  xmin=0, xmax=100,
  xtick={0,50,100},
  xticklabels={,,},
  ytick=\empty,
  title={Generators (left) vs. Real Validation Images (right)},
  title style={font=\scriptsize, yshift=-6pt},
  axis line style={axisgray!60, line width=0.4pt},
  tick style={axisgray!60, line width=0.3pt},
]
\draw[reflinecolor, densely dashed, line width=0.6pt]
  (rel axis cs:0.5,0) -- (rel axis cs:0.5,1);

\addplot+[fill=accentorange!85, draw=accentorange!60!black, line width=0.3pt]
coordinates { (\BARGen,1) (\RAEGen,2) (\OursGen,3) };
\addplot+[fill=slateblue!75, draw=slateblue!60!black, line width=0.3pt]
coordinates { (\BARVal,1) (\RAEVal,2) (\OursVal,3) };

\node[font=\tiny, anchor=east, text=black] at (axis cs:1, 1) {BAR-L$^\dagger$};
\node[font=\tiny, anchor=east, text=black] at (axis cs:1, 2) {RAE-XL$^\dagger$};
\node[font=\tiny\bfseries, anchor=east, text=accentorange!70!black] at (axis cs:1, 3) {Ours, {\ttiny pMF-H$^*$}};

\node[font=\tiny, anchor=west] at (axis cs:101, 1) {Val.};
\node[font=\tiny, anchor=west] at (axis cs:101, 2) {Val.};
\node[font=\tiny, anchor=west] at (axis cs:101, 3) {Val.};

\coordinate (BARg) at (axis cs:{\BARGen/2},1);
\coordinate (BARv) at (axis cs:{\BARGen + \BARVal/2},1);
\coordinate (RAEg) at (axis cs:{\RAEGen/2},2);
\coordinate (RAEv) at (axis cs:{\RAEGen + \RAEVal/2},2);
\coordinate (Oursg) at (axis cs:{\OursGen/2},3);
\coordinate (Oursv) at (axis cs:{\OursGen + \OursVal/2},3);
\end{axis}
\HPBarText{RAEg}{\RAEGen}{white}
\HPBarText{RAEv}{\RAEVal}{white}
\HPBarText{BARg}{\BARGen}{white}
\HPBarText{BARv}{\BARVal}{white}
\HPBarText{Oursg}{\OursGen}{white}
\HPBarText{Oursv}{\OursVal}{white}
\end{tikzpicture}
\end{minipage}
\vspace{-1.25em}
\caption{
\textbf{Human preference study.}
\textbf{Left}: Our post-trained 1-NFE models (\textcolor{accentorange!85!black}{\textbf{warm$^*$}}) are preferred over their base models.
\textbf{Right}: Our pMF-H$^*$ is the most preferred generator against real ImageNet validation images, but \emph{still} loses to real. This is consistent with Figure~\ref{fig:scatter_benchmark}: ImageNet generation is \emph{not} yet solved.
$^\dagger$: multi-step models; all post-trained models use FD-SIM.
Protocol in Appendix~\ref{sec:appendix_human_pref}.
}
\label{fig:human_pref}
\vspace{-1.em}
\end{figure}

\subsection{Text-Conditioned Generation}

Beyond class-conditioned generation models on ImageNet, we repurpose SD3.5~Medium~\cite{esser2024scaling}, a 2.5B-parameter MMDiT originally trained for multi-step latent-space denoising, into a 1-NFE text-to-image generator with \method (Fig.~\ref{fig:t2i}). This demonstrates that \method can extend beyond class-conditioned settings and scale to large text-conditioned models. A full comparison including a BLIP3o-Pretrain-Long-3M variant is in Figure~\ref{fig:t2i_appendix_full}; training details are in Appendix~\ref{sec:appendix_t2i}.
\input{sections/figures/fig_t2i}

\section{Conclusion}
\label{sec:conclusion}

\textit{Generation is, at its core, a distributional problem.} Over the years, however, the training of generative models has focused primarily on
sample-level losses, \eg, diffusion, flow matching, and adversarial objectives, while distributional distances, such as \frechet Distance, have lived only as
evaluators. This separation has been a matter of practicality rather than principle: FD has always been differentiable, yet reliable estimation requires
populations far beyond a training batch. Under these perspectives, our findings with \method are, in hindsight, a natural outcome: once population size and gradient size are decoupled, a distributional
distance can serve directly as a training loss.

We hope our work will make distribution-level post-training broadly applicable:
to other modalities, to settings where access to real data during post-training is scarce or restricted, and to paradigms beyond image generation.
More generally, once a distributional distance becomes optimizable at scale, a central design question shifts from \emph{how to optimize the distance} to
\emph{which representation space should define it}. Our work makes initial attempts by exploring several existing representations. 
Different representations induce different notions of visual similarity, and no single feature space should be expected to fully capture perceptual quality.
We hope this perspective will encourage future work on distribution-level objectives and representation-diverse evaluation for generative models.

\clearpage
\section*{Acknowledgments}
We are grateful to Tianhong Li, Tianyuan Zhang, Quankai Gao, Songlin Wei, and Rundi Wu for their helpful discussions and suggestions for this project.

The USC Physical Superintelligence Lab acknowledges generous supports from Toyota Research Institute, Dolby, Google DeepMind, Capital One, Nvidia, Bosch, NSF, and Qualcomm. Jiawei Yang is supported by the NVIDIA Graduate Fellowship. Yue Wang is also supported by a Powell Research Award.
\bibliographystyle{plain}
\bibliography{main}
\clearpage

\appendix
\numberwithin{table}{section}
\numberwithin{figure}{section}
\renewcommand{\thetable}{\thesection.\arabic{table}}
\setcounter{table}{0}
\renewcommand{\thefigure}{\thesection.\arabic{figure}}
\setcounter{figure}{0}

\begingroup
\hypersetup{linkcolor=black}
\begin{center}
{\LARGE\bfseries Appendix for\\[0.25em] Representation \frechet Loss for Visual Generation}
\end{center}
\vspace{0.8em}
\noindent{\bfseries Contents}
\begin{enumerate}[label=\Alph*., leftmargin=2em, itemsep=2pt, parsep=0pt]
  \item \hyperref[sec:appendix_design_attempts]{Additional Design Attempts and Observations}\hfill \apprange{sec:appendix_design_attempts}{sec:appendix_design_attempts_end}
  \item \hyperref[sec:appendix_impl]{Implementation and Evaluation Details}\hfill \apprange{sec:appendix_impl}{sec:appendix_impl_end}
  \item \hyperref[sec:appendix_human_pref]{Human Preference Study}\hfill \apprange{sec:appendix_human_pref}{sec:appendix_human_pref_end}
  \item \hyperref[sec:appendix_limitations]{Limitations and Broader Impact}\hfill \apprange{sec:appendix_limitations}{sec:appendix_limitations_end}
  \item \hyperref[sec:appendix_samples]{Additional Qualitative Samples}\hfill \apprange{sec:appendix_samples}{sec:appendix_samples_end}
  \item \hyperref[sec:appendix_detail]{Detailed Results}\hfill \apprange{sec:appendix_detail}{sec:appendix_detail_end}
  \item \hyperref[sec:t2i_prompts]{Text-to-Image Prompts}\hfill \apprange{sec:t2i_prompts}{sec:t2i_prompts_end}
\end{enumerate}
\vspace{0.8em}
\endgroup
\section{Additional Design Attempts and Observations}
\label{sec:appendix_design_attempts}

We summarize several exploratory attempts and observations that shaped the final design of \method. These should \emph{not} be read as negative results of our method, but rather as illustrations of what happens when a representation or objective is made too narrow. We hope these ``negative signals'' can serve as useful ``inverse gradients'' for future research.

\paragraph{Representation-coupled reward hacking.}
\emph{All} automatic objectives are proxies for the properties we ultimately care about. When the proxy is optimized directly, it can become misaligned with the underlying goal, a phenomenon often discussed through Goodhart's law~\cite{goodhart1984problems}. A closely related issue appears in reward-model optimization for large language models: reinforcement learning from human feedback methods learn a reward model from human preferences~\cite{christiano2017deep}, but over-optimizing that learned reward can reduce true preference quality~\cite{gao2023scaling}. Technically, \method is not an RL algorithm, but it faces an analogous proxy-optimization issue: reducing FD in a chosen feature space can be viewed as optimizing a distribution-level reward supplied by that representation model.

This proxy is useful, but it is incomplete. For example, optimizing Inception FD reliably improves the base generator and can bring pMF-B/16 to 0.81 FID; the resulting model is visually better than the base model. However, it is not necessarily perceptually stronger than some other generators whose FID is around 1.0--1.5. In this regime, continuing to optimize the same narrow representation further improves the chosen score without addressing the representation's blind spots. This is exactly the distinction exposed by \fdr{6}: Inception-only optimization greatly improves FID, while modern representation spaces reveal remaining quality gaps.

\begin{figure}
\centering
\includegraphics[width=\textwidth]{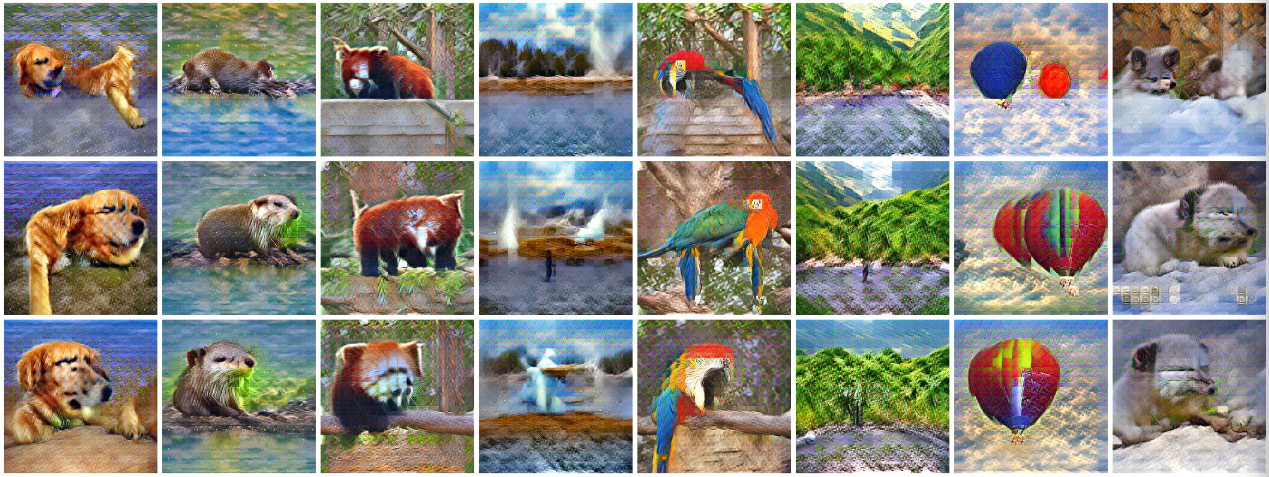}
\vspace{-0.6em}
\caption{
\textbf{A stress test of over-optimizing Inception-based metrics.}
We deliberately optimize Inception-based scores with a 100$\times$ larger learning rate. The resulting model attains 660 IS and 2.09 FID, but its samples exhibit clear artifacts and its \fdr{6} degrades to 50.66. This illustrates that Inception-based metrics can be pushed without improving perceptual quality, motivating representation-diverse evaluation and human preference checks.
}
\label{fig:appendix_reward_hacking}
\end{figure}

\paragraph{How much you can game FID and IS.}
The same point can be made more visibly by deliberately over-optimizing Inception-based metrics. In early experiments, we found that Inception Score (IS) can also be optimized with a queue-style estimator. As a focused stress test, we post-train pMF-B with a learning rate $10^{-4}$, which is $100{\times}$ larger than our default learning rate for pMF. Figure~\ref{fig:appendix_reward_hacking} shows the resulting samples: the model attains 660 IS and 2.09 FID, yet its \fdr{6} rises to 50.66, and the images exhibit clear artifacts and unnatural colors. We include these results only to show that Inception-based automatic metrics can be pushed to extremes, but on their own are not reliable objectives for visual quality. We omitted an extended discussion of IS optimization from the main paper for clarity. During early experimentation, we repeatedly observed similar variants that achieved exceptional FID and IS despite poor visual quality (\eg, FID between 0.9 and 2.1 with IS between 500 and 900), which ultimately motivated us to explore representation-diverse evaluation metrics.

\paragraph{The representation problem remains open.}
Even with multiple modern representation models, we believe some imperfections in generated images are still not captured by any representation model we tested. In this work, \fdr{6} uses six feature spaces spanning supervised, self-supervised, reconstructive, and vision-language objectives. This set is substantially more diverse than FID alone, and it reveals many gaps that Inception misses. Still, six representations are unlikely to exhaust perceptual quality. Every representation discards some information, and every automatic metric inherits the blind spots of its feature space. In this sense, finding representations that capture perceptual quality is an open-ended problem rather than a finite checklist.

We view \fdr{k} as a practical step toward reducing single-model bias, not as a final solution. This is also why we conduct a human preference study. When one-step post-trained models score better than strong multi-step systems such as RAE under \fdr{6}, we need to check whether they genuinely look better to humans, or whether we have merely found a broader but still hackable automatic metric.

\label{sec:appendix_design_attempts_end}

\section{Implementation and Evaluation Details}
\label{sec:appendix_impl}

\subsection{Representation Models for \fdr{6}}
\label{sec:appendix_repr_models}

Table~\ref{tab:repr_models} summarizes the representation models used to compute \fdr{6}. These span supervised, self-supervised, and vision-language training objectives across both CNN and ViT architectures.

\begin{table*}[h]
\begin{center}
\caption{\textbf{Representation models used in \fdr{6}.}}
\label{tab:repr_models}
\vspace{.3em}
\tablestyle{3pt}{1.03}
\begin{tabular}{l l c c c c}
model & timm identifier & arch. & dim & objective & pooling \\
\shline
Inception-v3~\cite{szegedy2016rethinking} & \texttt{inception\_v3} (torch-fidelity) & CNN & 2048 & supervised & global avg pool \\
ConvNeXt-v2~\cite{liu2022convnet} & \texttt{convnextv2\_base.fcmae\_ft\_in22k\_in1k} & CNN & 1024 & self-supervised & global avg pool \\
MAE~\cite{he2022masked} & \texttt{vit\_large\_patch16\_224.mae} & ViT & 1024 & reconstructive & CLS token \\
DINOv2~\cite{oquab2024dinov2} & \texttt{vit\_large\_patch14\_dinov2.lvd142m} & ViT & 1024 & contrastive & CLS token \\
SigLIP2~\cite{tschannen2025siglip} & \texttt{vit\_so400m\_patch16\_siglip\_256.v2\_webli} & ViT & 1152 & vision-language & CLS token \\
CLIP~\cite{radford2021learning} & \texttt{vit\_large\_patch14\_clip\_224.openai} & ViT & 1024 & vision-language & CLS token \\
\end{tabular}%
\end{center}
\end{table*}

For CNN-based models, features are extracted after the final spatial pooling layer. For ViT-based models, we use the CLS token from the final layer. All representation models are frozen during training.

\subsection{Configurations}
\label{sec:appendix_configs}

Table~\ref{tab:configs} summarizes the configurations used for ImageNet class-conditional post-training across the three generator families (pMF, iMF, JiT).

\begin{table}[h]
\centering
\tablestyle{5pt}{1.05}
\begin{tabular}{l|ccc}
 & \textbf{pMF}~\cite{lu2026one} & \textbf{iMF}~\cite{geng2025improved} & \textbf{JiT}~\cite{li2025back} \\
\shline
\rowcolor[gray]{0.9}\multicolumn{4}{l}{\textbf{base model}} \\
space & pixel & latent (SD-VAE) & pixel \\
sizes (256px) & B, L, H & B, L, XL & B, L, H \\
sizes (512px) & B, L, H & \multicolumn{2}{c}{not used} \\
patch size & \multicolumn{3}{c}{16 (256px), 32 (512px)} \\
initialization & \multicolumn{3}{c}{official pre-trained weights} \\
\hline
\rowcolor[gray]{0.9}\multicolumn{4}{l}{\textbf{representation models for \fdr{}}} \\
FD-Incep.\ & \multicolumn{3}{c}{Inception-v3~\cite{szegedy2016rethinking}} \\
FD-SIM & \multicolumn{3}{c}{SigLIP2~\cite{tschannen2025siglip} + Inception-v3~\cite{szegedy2016rethinking} + MAE~\cite{he2022masked}} \\
input resolution & \multicolumn{3}{c}{$224$ (SigLIP2), $299$ (Incep.), $224$ (MAE)} \\
normalization $c$ & \multicolumn{3}{c}{$0.01$} \\
matrix square root & \multicolumn{3}{c}{\texttt{torch.linalg.eigvalsh}} \\
\hline
\rowcolor[gray]{0.9}\multicolumn{4}{l}{\textbf{statistics estimator}} \\
default estimator & \multicolumn{3}{c}{EMA, $\beta{=}0.999$} \\
queue size (ablation) & \multicolumn{3}{c}{see Table~\ref{tab:queue_size}} \\
warm-start samples & \multicolumn{3}{c}{50k generated from base model} \\
\hline
\rowcolor[gray]{0.9}\multicolumn{4}{l}{\textbf{training}} \\
epochs (Tables~\ref{tab:queue_size}, \ref{tab:ema_beta}, \ref{tab:backbone}, \ref{tab:repurpose}) & \multicolumn{3}{c}{50} \\
epochs (Tables~\ref{tab:all_fd_models}, \ref{tab:system}) & \multicolumn{3}{c}{100} \\
optimizer & \multicolumn{3}{c}{AdamW~\cite{Loshchilov2019}, $\beta_1, \beta_2 {=} 0.9, 0.95$} \\
weight decay & \multicolumn{3}{c}{0} \\
learning rate & $1\text{e-}6$ & $1\text{e-}6$ & $1\text{e-}5$ \\
lr schedule & \multicolumn{3}{c}{cosine} \\
warmup epochs & \multicolumn{3}{c}{5} \\
global batch size & \multicolumn{3}{c}{1024} \\
precision & \multicolumn{3}{c}{bf16} \\
gradient clipping & \multicolumn{3}{c}{none} \\
dropout & \multicolumn{3}{c}{0} \\
model-weight EMA & \multicolumn{3}{c}{none (online weights)} \\
augmentation & \multicolumn{3}{c}{center crop, horizontal flip} \\
\hline
\rowcolor[gray]{0.9}\multicolumn{4}{l}{\textbf{sampling / evaluation}} \\
NFE & \multicolumn{3}{c}{1} \\
CFG at inference & default (CFG-distilled) & default (CFG-distilled) & $1.0$ \\
\end{tabular}
\vspace{.2em}
\caption{\textbf{Configurations for ImageNet class-conditional post-training} with \method. All settings are shared across ablation and final runs unless noted. Base models are used as released; only post-training differs from the source papers.}
\label{tab:configs}
\end{table}

\subsection{Training Details}
\label{sec:appendix_training}

\paragraph{Initialization.}
Before training begins, we generate images \emph{on-the-fly} from the base model to initialize the statistics estimators. For the EMA variant, we generate 50k images and compute the initial feature moments $(\boldsymbol{\mu}^{(0)}_g, \mathbf{M}^{(0)}_g)$ from these samples. For the queue variant, we similarly generate $N$ images (where $N$ is the queue size) and fill the queue with their features. In both cases, this provides a warm start so that the FD estimate is meaningful from the first training step.

\paragraph{Matrix square root.}
Computing FD (Eq.~\ref{eq:fd}) requires the matrix square root $(\boldsymbol{\Sigma}_r \boldsymbol{\Sigma}_g)^{1/2}$. We precompute $\boldsymbol{\Sigma}_r^{1/2}$ via eigendecomposition and then compute the trace term efficiently using \texttt{torch.linalg.eigvalsh} on the symmetric product $\boldsymbol{\Sigma}_r^{1/2} \boldsymbol{\Sigma}_g \boldsymbol{\Sigma}_r^{1/2}$, avoiding an explicit matrix square root at each training step. In early exploration, we compared \texttt{torch.linalg.eigvals} and \texttt{torch.linalg.eigvalsh} and found them to perform similarly, with the latter being significantly faster; we therefore adopt it.

\subsection{Text-to-Image Post-Training}
\label{sec:appendix_t2i}

We take the MMDiT transformer from Stable Diffusion~3.5 Medium~\cite{esser2024scaling} (2.5B parameters) and post-train it with \method for one-step generation at $256{\times}256$ resolution.
The SD3.5 VAE tokenizer is used unchanged.
We use the SIM representation set (SigLIP2+Inception+MAE).
Reference statistics $(\boldsymbol{\mu}_r, \boldsymbol{\Sigma}_r)$ are pre-computed from all real images in each set (3M for variant (i), 60k for variant (ii)).
EMA feature statistics are warm-started with 50k images generated from the base model before training begins.

Training uses a cosine learning rate schedule with peak $\text{lr}{=}10^{-5}$ and 2{,}500 warmup steps, for 15{,}000 total steps.
The global batch size is 1024; each step generates one image per caption with one denoising step (no classifier-free guidance, $\text{CFG}{=}1$).
EMA statistics are tracked with $\beta{=}0.999$ and the eigenvalue-based matrix square root (\texttt{eigvalsh}) is used.

We train two variants that differ only in the caption and image sources:
\textbf{(i)}~BLIP3o-Pretrain-Long-3M~\cite{chen2025blip3}, a 3M subset randomly sampled from the original 30M caption-image web dataset, paired with realistic photographic images; and
\textbf{(ii)}~BLIP3o-GPT4o-60k~\cite{chen2025blip3}, a 60k curated set whose images are distilled from GPT-4o and exhibit a stylized, illustration-leaning aesthetic.
All other hyperparameters are identical between the two runs.

\begin{table}[h]
\centering
\tablestyle{5pt}{1.05}
\begin{tabular}{l|c}
 & \textbf{SD3.5 Medium~\cite{esser2024scaling}} \\
\shline
\rowcolor[gray]{0.9}\multicolumn{2}{l}{\textbf{base model}} \\
architecture & MMDiT, 2.5B params \\
resolution & $256{\times}256$ \\
tokenizer & SD3.5 VAE  \\
\hline
\rowcolor[gray]{0.9}\multicolumn{2}{l}{\textbf{representation models for \fdr{}}} \\
FD-SIM & SigLIP2 + Inception-v3 + MAE \\
\hline
\rowcolor[gray]{0.9}\multicolumn{2}{l}{\textbf{statistics estimator}} \\
estimator & EMA, $\beta{=}0.999$ \\
warm-start & 50k samples from base \\
reference stats $(\boldsymbol{\mu}_r, \boldsymbol{\Sigma}_r)$ & computed from all images in each set (3M / 60k) \\
\hline
\rowcolor[gray]{0.9}\multicolumn{2}{l}{\textbf{training}} \\
total steps & 15{,}000 \\
warmup steps & 2{,}500 \\
optimizer & AdamW, $\beta_1, \beta_2 {=} 0.9, 0.95$, wd $=0$ \\
peak learning rate & $1\text{e-}5$ \\
lr schedule & cosine \\
global batch size & 1024 \\
precision & bf16 \\
gradient clipping & none \\
dropout & 0 \\
model-weight EMA & none \\
\hline
\rowcolor[gray]{0.9}\multicolumn{2}{l}{\textbf{sampling}} \\
NFE & 1 \\
CFG & $1.0$ \\
\hline
\rowcolor[gray]{0.9}\multicolumn{2}{l}{\textbf{caption sets (two variants)}} \\
(i) photographic & BLIP3o-Pretrain-Long-3M~\cite{chen2025blip3} \\
(ii) stylized & BLIP3o-GPT4o-60k~\cite{chen2025blip3} (GPT-4o distilled) \\
\end{tabular}
\vspace{.2em}
\caption{\textbf{Configurations for text-to-image post-training} of SD3.5 Medium with \method. The two variants differ only in the caption source.}
\label{tab:configs_t2i}
\end{table}

\label{sec:appendix_t2i_end}

\subsection{Evaluation Protocol}
\label{sec:appendix_eval}

For all methods, we sample 50,000 images from the official checkpoints using the official code, primarily relying on coding agents such as Claude Code~\cite{anthropic2025claudecode} and Codex~\cite{openai2025codex} to follow the instructions provided in each codebase. We then manually check if the sampled images are correct and run the evaluation on the sampled images ourselves. Class-conditional models sample uniformly across all 1,000 classes (50 images per class). Reference statistics $(\boldsymbol{\mu}_r, \boldsymbol{\Sigma}_r)$ are computed once from the full ImageNet training set.

\label{sec:appendix_eval_end}
\label{sec:appendix_impl_end}

\section{Human Preference Study}
\label{sec:appendix_human_pref}

We conduct a pairwise human preference study using anonymized $3{\times}3$ image grids.
For each trial, voters choose the grid with higher visual fidelity, with an additional tie option.
The left--right order is randomized independently for every trial, and model identities are hidden.
We evaluate two settings: (i)~\emph{Post-trained vs. Base}, where each \method post-trained model is compared with its corresponding base model using matched initial noise, so the two grids are directly comparable; and (ii)~\emph{Generator vs. Real}, where a generator is compared against ImageNet validation images.

We collect $2{,}929$ valid votes from $17$ participants after filtering incomplete or invalid responses.
For reporting, ties are split evenly between the two sides: if $W$, $T$, and $L$ denote win, tie, and loss rates for \method, its preference score is $W + T/2$.
Figure~\ref{fig:human_pref} reports aggregated preference scores.

\begin{figure}[t]
\centering
\includegraphics[width=\textwidth]{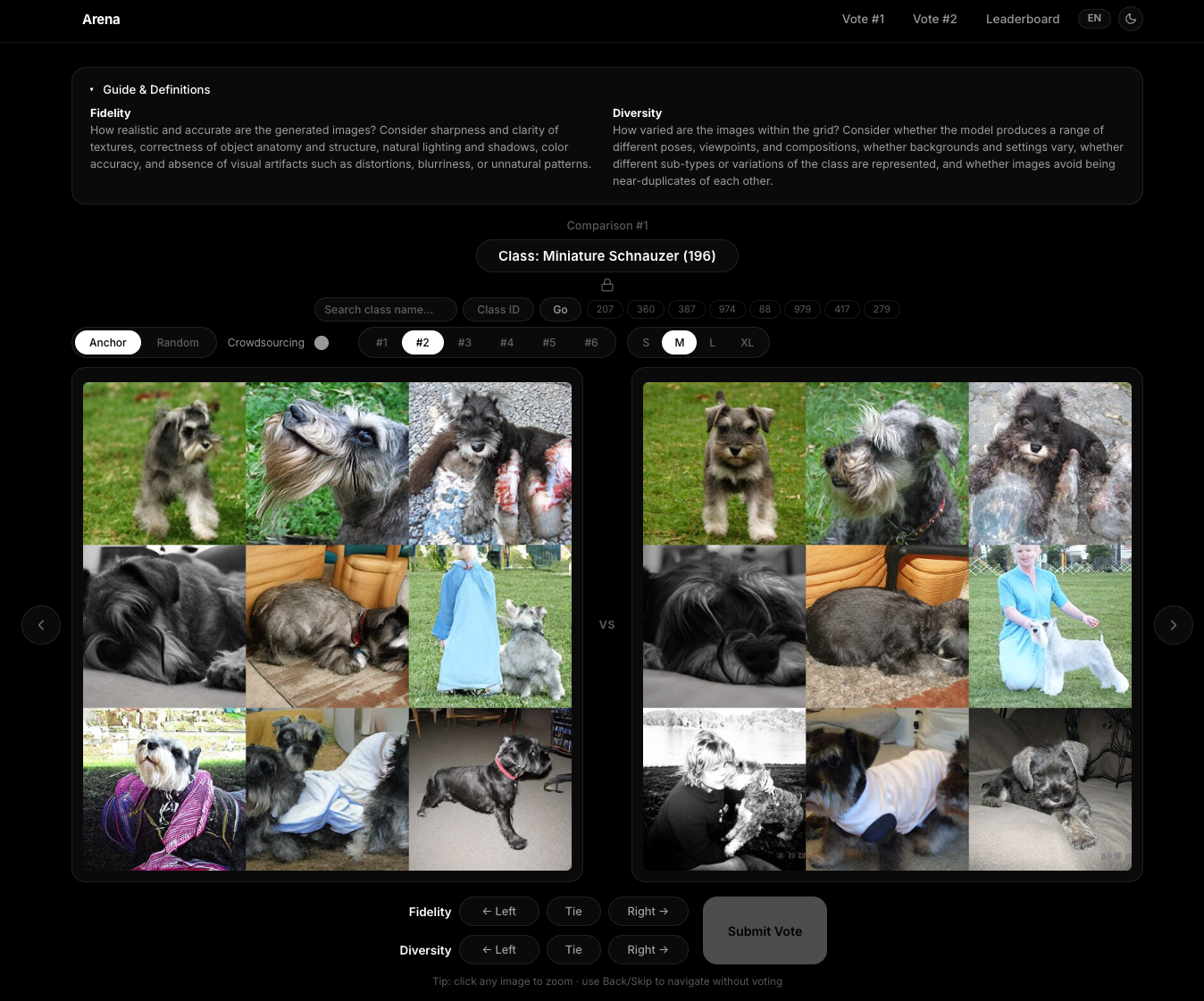}
\vspace{-0.6em}
\caption{
Screenshot of the voting page. Voters see two anonymized $3{\times}3$ grids of samples for the same ImageNet class and pick \emph{Left}, \emph{Tie}, or \emph{Right} for fidelity.
}
\label{fig:arena_screenshot}
\end{figure}

\paragraph{Interface.}
The voting interface is shown in Figure~\ref{fig:arena_screenshot}.
Each trial shows two side-by-side grids of uncurated samples from the same ImageNet class, sampled uniformly from the $1000$ classes.
The voter selects \emph{Left}, \emph{Tie}, or \emph{Right}.
A \emph{Skip} arrow advances to a new pair without recording a vote, and a \emph{Back} arrow revisits the immediately preceding pair.
Images can be clicked for closer inspection.

\paragraph{Sampling.}
For each model, we sample 50{,}000 images in total, with 50 images per class.
For each class, images are grouped into $3{\times}3$ grids; when the number of images is insufficient for an integer number of grids, we sample with replacement.
All generators are sampled using their best-FID inference settings.
\emph{Post-trained vs. Base} trials are sampled uniformly among the three base/post-trained pairs, and \emph{Generator vs. Real} trials are sampled uniformly among the three generator/validation pairs.

\label{sec:appendix_human_pref_end}

\section{Limitations and Broader Impact}
\label{sec:appendix_limitations}

\paragraph{Limitations.}
Our study focuses on image generation, with ImageNet serving as the primary controlled benchmark and text-conditioned generation providing an additional demonstration. As with any distribution-level objective or automatic metric, the behavior of \method depends on the choice of representation spaces and reference statistics; different domains may benefit from different feature sets or weighting schemes. \method is designed as a post-training objective that complements existing generator training recipes and evaluation protocols. Future work can explore broader data distributions, additional modalities, higher-resolution settings, and adaptive or learned representation sets for distribution-level optimization.

\paragraph{Broader impact.}
Our work improves the quality of image generators, which carries dual-use risks common to all generative modeling research. Higher-quality generators could be misused to generate disinformation or deceptive content. We believe that the diagnostic value of our work, showing that FID alone is an insufficient quality measure, contributes positively to the field's ability to evaluate and understand generative models. We do not release any new datasets; all experiments use publicly available models and data.

\label{sec:appendix_limitations_end}

\section{Additional Qualitative Samples}
\label{sec:appendix_samples}

We provide uncurated paired samples for two generators from Table~\ref{tab:system}. First, we show the one-step pMF-H/16 base model alongside its post-trained counterpart using \method (SIM). Second, we show JiT-H/16 with 200 NFE (50 steps $\times$ 2 (Heun) $\times$ 2 (CFG)) alongside its \method post-trained version using only 1 NFE. Each pair uses the same initial noise for direct comparison. All samples are uncurated.

\newcommand{\hhs}{\hspace{-0.001em}}
\newcommand{\vvs}{\vspace{-.1em}}

\newcommand{\sysbasedir}{figures/appendix_system/pMF_H_base}
\newcommand{\syssimdir}{figures/appendix_system/pMF_H_fd_sim}
\newcommand{\systilewidth}{0.14\linewidth}
\newcommand{\samplegridgap}{1pt}

\newcommand{\syscaptiontext}{
\textit{Uncurated} paired samples on ImageNet 256$\times$256.
Each class shows the \textbf{base model} pMF-H/16~\cite{lu2026one} (left) and the \textbf{post-trained} pMF-H/16 with \method (SIM) (right), using the \textit{same initial noise}.
SIM: SigLIP+Inception+MAE.
}

\newcommand{\syscoltitles}{%
\begin{minipage}[t]{0.49\linewidth}\centering
{\scriptsize\textcolor{gray}{pMF-H/16 (base model, 1 NFE)}}%
\end{minipage}\hfill
\begin{minipage}[t]{0.49\linewidth}\centering
{\scriptsize\textcolor{gray}{pMF-H/16 + \method (1 NFE)}}%
\end{minipage}\par\vspace{1pt}%
}

\newcommand{\addclasspair}[2]{%
\begin{minipage}[t]{0.49\linewidth}
\centering
\raisebox{-\systilewidth}[0pt][0pt]{\begin{minipage}[t]{0.28\linewidth}%
\includegraphics[width=\linewidth]{\sysbasedir/cls#1_r00.jpg}\vvs\\
\includegraphics[width=\linewidth]{\sysbasedir/cls#1_r01.jpg}%
\end{minipage}}%
\hhs%
\begin{minipage}[t]{0.70\linewidth}%
\includegraphics[width=0.2\linewidth]{\sysbasedir/cls#1_r02.jpg}\hhs
\includegraphics[width=0.2\linewidth]{\sysbasedir/cls#1_r03.jpg}\hhs
\includegraphics[width=0.2\linewidth]{\sysbasedir/cls#1_r04.jpg}\hhs
\includegraphics[width=0.2\linewidth]{\sysbasedir/cls#1_r05.jpg}\hhs
\includegraphics[width=0.2\linewidth]{\sysbasedir/cls#1_r06.jpg}\vvs
\\
\includegraphics[width=0.2\linewidth]{\sysbasedir/cls#1_r07.jpg}\hhs
\includegraphics[width=0.2\linewidth]{\sysbasedir/cls#1_r08.jpg}\hhs
\includegraphics[width=0.2\linewidth]{\sysbasedir/cls#1_r09.jpg}\hhs
\includegraphics[width=0.2\linewidth]{\sysbasedir/cls#1_r10.jpg}\hhs
\includegraphics[width=0.2\linewidth]{\sysbasedir/cls#1_r11.jpg}\vvs
\\
\includegraphics[width=0.2\linewidth]{\sysbasedir/cls#1_r12.jpg}\hhs
\includegraphics[width=0.2\linewidth]{\sysbasedir/cls#1_r13.jpg}\hhs
\includegraphics[width=0.2\linewidth]{\sysbasedir/cls#1_r14.jpg}\hhs
\includegraphics[width=0.2\linewidth]{\sysbasedir/cls#1_r15.jpg}\hhs
\includegraphics[width=0.2\linewidth]{\sysbasedir/cls#1_r16.jpg}\vvs
\\
\includegraphics[width=0.2\linewidth]{\sysbasedir/cls#1_r17.jpg}\hhs
\includegraphics[width=0.2\linewidth]{\sysbasedir/cls#1_r18.jpg}\hhs
\includegraphics[width=0.2\linewidth]{\sysbasedir/cls#1_r19.jpg}\hhs
\includegraphics[width=0.2\linewidth]{\sysbasedir/cls#1_r20.jpg}\hhs
\includegraphics[width=0.2\linewidth]{\sysbasedir/cls#1_r21.jpg}%
\end{minipage}%
\par
{\scriptsize {class #1}: #2}
\vspace{.2em}
\end{minipage}
\hfill
\begin{minipage}[t]{0.49\linewidth}
\centering
\raisebox{-\systilewidth}[0pt][0pt]{\begin{minipage}[t]{0.28\linewidth}%
\includegraphics[width=\linewidth]{\syssimdir/cls#1_r00.jpg}\vvs\\
\includegraphics[width=\linewidth]{\syssimdir/cls#1_r01.jpg}%
\end{minipage}}%
\hhs%
\begin{minipage}[t]{0.70\linewidth}%
\includegraphics[width=0.2\linewidth]{\syssimdir/cls#1_r02.jpg}\hhs
\includegraphics[width=0.2\linewidth]{\syssimdir/cls#1_r03.jpg}\hhs
\includegraphics[width=0.2\linewidth]{\syssimdir/cls#1_r04.jpg}\hhs
\includegraphics[width=0.2\linewidth]{\syssimdir/cls#1_r05.jpg}\hhs
\includegraphics[width=0.2\linewidth]{\syssimdir/cls#1_r06.jpg}\vvs
\\
\includegraphics[width=0.2\linewidth]{\syssimdir/cls#1_r07.jpg}\hhs
\includegraphics[width=0.2\linewidth]{\syssimdir/cls#1_r08.jpg}\hhs
\includegraphics[width=0.2\linewidth]{\syssimdir/cls#1_r09.jpg}\hhs
\includegraphics[width=0.2\linewidth]{\syssimdir/cls#1_r10.jpg}\hhs
\includegraphics[width=0.2\linewidth]{\syssimdir/cls#1_r11.jpg}\vvs
\\
\includegraphics[width=0.2\linewidth]{\syssimdir/cls#1_r12.jpg}\hhs
\includegraphics[width=0.2\linewidth]{\syssimdir/cls#1_r13.jpg}\hhs
\includegraphics[width=0.2\linewidth]{\syssimdir/cls#1_r14.jpg}\hhs
\includegraphics[width=0.2\linewidth]{\syssimdir/cls#1_r15.jpg}\hhs
\includegraphics[width=0.2\linewidth]{\syssimdir/cls#1_r16.jpg}\vvs
\\
\includegraphics[width=0.2\linewidth]{\syssimdir/cls#1_r17.jpg}\hhs
\includegraphics[width=0.2\linewidth]{\syssimdir/cls#1_r18.jpg}\hhs
\includegraphics[width=0.2\linewidth]{\syssimdir/cls#1_r19.jpg}\hhs
\includegraphics[width=0.2\linewidth]{\syssimdir/cls#1_r20.jpg}\hhs
\includegraphics[width=0.2\linewidth]{\syssimdir/cls#1_r21.jpg}%
\end{minipage}%
\par
{\scriptsize {class #1}: #2}
\vspace{.2em}
\end{minipage}
}

\begin{figure*}[p]
\centering
\syscoltitles
\addclasspair{0088}{macaw}
\par\vspace{\samplegridgap}
\addclasspair{0117}{chambered nautilus, pearly nautilus}
\par\vspace{\samplegridgap}
\addclasspair{0207}{golden retriever}
\par\vspace{\samplegridgap}
\addclasspair{0279}{Arctic fox, white fox}
\par\vspace{\samplegridgap}
\addclasspair{0288}{leopard, Panthera pardus}
\par\vspace{\samplegridgap}
\caption{\syscaptiontext}
\label{fig:appendix_samples}
\end{figure*}
\clearpage

\begin{figure*}[p]
\centering
\syscoltitles
\addclasspair{0349}{bighorn, bighorn sheep}
\par\vspace{\samplegridgap}
\addclasspair{0387}{lesser panda, red panda}
\par\vspace{\samplegridgap}
\addclasspair{0425}{barn}
\par\vspace{\samplegridgap}
\addclasspair{0453}{bookcase}
\par\vspace{\samplegridgap}
\addclasspair{0661}{Model T}
\par\vspace{\samplegridgap}
\caption{\syscaptiontext (cont.)}
\end{figure*}
\clearpage

\begin{figure*}[p]
\centering
\syscoltitles
\addclasspair{0718}{pier}
\par\vspace{\samplegridgap}
\addclasspair{0725}{pitcher, ewer}
\par\vspace{\samplegridgap}
\addclasspair{0757}{recreational vehicle, RV}
\par\vspace{\samplegridgap}
\addclasspair{0829}{streetcar, tram}
\par\vspace{\samplegridgap}
\addclasspair{0873}{triumphal arch}
\par\vspace{\samplegridgap}
\caption{\syscaptiontext (cont.)}
\end{figure*}
\clearpage

\newcommand{\jitbasedir}{figures/appendix_system/JiT_H_base}
\newcommand{\jitsimdir}{figures/appendix_system/JiT_H_fd_sim}

\newcommand{\jitcaptiontext}{
\textit{Uncurated} paired samples on ImageNet 256$\times$256.
Each class shows the \textbf{base model} JiT-H/16~\cite{li2025back} with 200 NFE (50 steps $\times$ 2 (Heun) $\times$ 2 (CFG)) (left) and the \textbf{post-trained} JiT-H/16 with \method (SIM), 1 NFE (right), using the \textit{same initial noise}.
SIM: SigLIP+Inception+MAE.
}

\newcommand{\jitcoltitles}{%
\begin{minipage}[t]{0.49\linewidth}\centering
{\scriptsize\textcolor{gray}{JiT-H/16 (base model, 200 NFE)}}%
\end{minipage}\hfill
\begin{minipage}[t]{0.49\linewidth}\centering
{\scriptsize\textcolor{gray}{JiT-H/16 + \method (1 NFE)}}%
\end{minipage}\par\vspace{1pt}%
}

\newcommand{\addjitpair}[2]{%
\begin{minipage}[t]{0.49\linewidth}
\centering
\raisebox{-\systilewidth}[0pt][0pt]{\begin{minipage}[t]{0.28\linewidth}%
\includegraphics[width=\linewidth]{\jitbasedir/cls#1_r00.jpg}\vvs\\
\includegraphics[width=\linewidth]{\jitbasedir/cls#1_r01.jpg}%
\end{minipage}}%
\hhs%
\begin{minipage}[t]{0.70\linewidth}%
\includegraphics[width=0.2\linewidth]{\jitbasedir/cls#1_r02.jpg}\hhs
\includegraphics[width=0.2\linewidth]{\jitbasedir/cls#1_r03.jpg}\hhs
\includegraphics[width=0.2\linewidth]{\jitbasedir/cls#1_r04.jpg}\hhs
\includegraphics[width=0.2\linewidth]{\jitbasedir/cls#1_r05.jpg}\hhs
\includegraphics[width=0.2\linewidth]{\jitbasedir/cls#1_r06.jpg}\vvs
\\
\includegraphics[width=0.2\linewidth]{\jitbasedir/cls#1_r07.jpg}\hhs
\includegraphics[width=0.2\linewidth]{\jitbasedir/cls#1_r08.jpg}\hhs
\includegraphics[width=0.2\linewidth]{\jitbasedir/cls#1_r09.jpg}\hhs
\includegraphics[width=0.2\linewidth]{\jitbasedir/cls#1_r10.jpg}\hhs
\includegraphics[width=0.2\linewidth]{\jitbasedir/cls#1_r11.jpg}\vvs
\\
\includegraphics[width=0.2\linewidth]{\jitbasedir/cls#1_r12.jpg}\hhs
\includegraphics[width=0.2\linewidth]{\jitbasedir/cls#1_r13.jpg}\hhs
\includegraphics[width=0.2\linewidth]{\jitbasedir/cls#1_r14.jpg}\hhs
\includegraphics[width=0.2\linewidth]{\jitbasedir/cls#1_r15.jpg}\hhs
\includegraphics[width=0.2\linewidth]{\jitbasedir/cls#1_r16.jpg}\vvs
\\
\includegraphics[width=0.2\linewidth]{\jitbasedir/cls#1_r17.jpg}\hhs
\includegraphics[width=0.2\linewidth]{\jitbasedir/cls#1_r18.jpg}\hhs
\includegraphics[width=0.2\linewidth]{\jitbasedir/cls#1_r19.jpg}\hhs
\includegraphics[width=0.2\linewidth]{\jitbasedir/cls#1_r20.jpg}\hhs
\includegraphics[width=0.2\linewidth]{\jitbasedir/cls#1_r21.jpg}%
\end{minipage}%
\par
{\scriptsize {class #1}: #2}
\vspace{.2em}
\end{minipage}
\hfill
\begin{minipage}[t]{0.49\linewidth}
\centering
\raisebox{-\systilewidth}[0pt][0pt]{\begin{minipage}[t]{0.28\linewidth}%
\includegraphics[width=\linewidth]{\jitsimdir/cls#1_r00.jpg}\vvs\\
\includegraphics[width=\linewidth]{\jitsimdir/cls#1_r01.jpg}%
\end{minipage}}%
\hhs%
\begin{minipage}[t]{0.70\linewidth}%
\includegraphics[width=0.2\linewidth]{\jitsimdir/cls#1_r02.jpg}\hhs
\includegraphics[width=0.2\linewidth]{\jitsimdir/cls#1_r03.jpg}\hhs
\includegraphics[width=0.2\linewidth]{\jitsimdir/cls#1_r04.jpg}\hhs
\includegraphics[width=0.2\linewidth]{\jitsimdir/cls#1_r05.jpg}\hhs
\includegraphics[width=0.2\linewidth]{\jitsimdir/cls#1_r06.jpg}\vvs
\\
\includegraphics[width=0.2\linewidth]{\jitsimdir/cls#1_r07.jpg}\hhs
\includegraphics[width=0.2\linewidth]{\jitsimdir/cls#1_r08.jpg}\hhs
\includegraphics[width=0.2\linewidth]{\jitsimdir/cls#1_r09.jpg}\hhs
\includegraphics[width=0.2\linewidth]{\jitsimdir/cls#1_r10.jpg}\hhs
\includegraphics[width=0.2\linewidth]{\jitsimdir/cls#1_r11.jpg}\vvs
\\
\includegraphics[width=0.2\linewidth]{\jitsimdir/cls#1_r12.jpg}\hhs
\includegraphics[width=0.2\linewidth]{\jitsimdir/cls#1_r13.jpg}\hhs
\includegraphics[width=0.2\linewidth]{\jitsimdir/cls#1_r14.jpg}\hhs
\includegraphics[width=0.2\linewidth]{\jitsimdir/cls#1_r15.jpg}\hhs
\includegraphics[width=0.2\linewidth]{\jitsimdir/cls#1_r16.jpg}\vvs
\\
\includegraphics[width=0.2\linewidth]{\jitsimdir/cls#1_r17.jpg}\hhs
\includegraphics[width=0.2\linewidth]{\jitsimdir/cls#1_r18.jpg}\hhs
\includegraphics[width=0.2\linewidth]{\jitsimdir/cls#1_r19.jpg}\hhs
\includegraphics[width=0.2\linewidth]{\jitsimdir/cls#1_r20.jpg}\hhs
\includegraphics[width=0.2\linewidth]{\jitsimdir/cls#1_r21.jpg}%
\end{minipage}%
\par
{\scriptsize {class #1}: #2}
\vspace{.2em}
\end{minipage}
}

\begin{figure*}[p]
\centering
\jitcoltitles
\addjitpair{0207}{golden retriever}
\par\vspace{\samplegridgap}
\addjitpair{0279}{Arctic fox, white fox}
\par\vspace{\samplegridgap}
\addjitpair{0288}{leopard, Panthera pardus}
\par\vspace{\samplegridgap}
\addjitpair{0387}{lesser panda, red panda}
\par\vspace{\samplegridgap}
\addjitpair{0661}{Model T}
\par\vspace{\samplegridgap}
\caption{\jitcaptiontext}
\label{fig:appendix_jit_samples}
\end{figure*}
\clearpage


\label{sec:appendix_samples_end}

\clearpage
\section{Detailed Results}
\label{sec:appendix_detail}

This appendix provides per-representation breakdowns for all experiments, in both FDr (ratio to validation set, Eq.~\ref{eq:fd_ratio}) and raw FD (\frechet Distance).
\fdr{6} is the arithmetic mean of FDr over six representation spaces (Incep., ConvNeXt, DINOv2, MAE, SigLIP, CLIP).
FDr-CLIP is additionally reported as a held-out evaluator.

\begin{table*}[h]
\begin{center}
\caption{
\textbf{Per-representation FDr for population size ablation} (Tabs.~\ref{tab:queue_size} and~\ref{tab:ema_beta}).
pMF-B/16 post-trained for 50 epochs with FD-Inception.
Default setting in \baselinelegend{Gray}.
$^\dagger$Statistics from current batch only.
}
\label{tab:detail_population}
\vspace{.3em}
\tablestyle{3pt}{1.03}
\begin{tabular}{l | c c c c c | c | c c c}
setting & Incep. & ConvNeXt & DINOv2 & MAE & SigLIP & FDr-CLIP & FID$\downarrow$ & IS$\uparrow$ & \fdr{6}$\downarrow$ \\
\shline
\multicolumn{10}{l}{\hspace{-.5em} \textit{Queue size}} \\
\gc{Base} & \gc{1.98} & \gc{1.93} & \gc{10.13} & \gc{13.81} & \gc{31.03} & \gc{23.30} & \gc{3.31} & \gc{254.6} & \gc{13.70} \\
\gc{0k$^\dagger$} & \gc{2.29} & \gc{3.99} & \gc{13.74} & \gc{16.73} & \gc{35.49} & \gc{30.15} & \gc{3.84} & \gc{250.9} & \gc{17.06} \\
5k & 0.62 & 1.41 & 8.73 & 9.54 & 28.15 & 22.89 & 1.05 & 280.0 & 11.89 \\
10k & 0.56 & 1.36 & 8.35 & 9.02 & 28.08 & 22.87 & 0.93 & 283.9 & 11.71 \\
\baseline{50k} & \baseline{0.53} & \baseline{1.50} & \baseline{7.83} & \baseline{8.28} & \baseline{26.12} & \baseline{21.20} & \baseline{0.89} & \baseline{288.3} & \baseline{10.91} \\
100k & 0.56 & 1.60 & 7.84 & 9.06 & 26.63 & 21.22 & 0.93 & 288.8 & 11.15 \\
500k & 0.72 & 2.17 & 9.49 & 19.57 & 40.11 & 33.92 & 1.22 & 294.4 & 17.67 \\
\hline
\multicolumn{10}{l}{\hspace{-.5em} \textit{EMA decay rate ($\beta$)}} \\
\gc{Base} & \gc{1.98} & \gc{1.93} & \gc{10.13} & \gc{13.81} & \gc{31.03} & \gc{23.30} & \gc{3.31} & \gc{254.6} & \gc{13.70} \\
\gc{0.0$^\dagger$} & \gc{2.29} & \gc{3.99} & \gc{13.74} & \gc{16.73} & \gc{35.49} & \gc{30.15} & \gc{3.84} & \gc{250.9} & \gc{17.06} \\
0.9 & 0.58 & 1.50 & 8.28 & 8.74 & 26.21 & 21.84 & 0.98 & 283.6 & 11.19 \\
0.99 & 0.50 & 1.34 & 7.57 & 8.51 & 25.50 & 20.98 & 0.84 & 291.8 & 10.74 \\
\baseline{0.999} & \baseline{0.48} & \baseline{1.26} & \baseline{7.52} & \baseline{8.51} & \baseline{26.02} & \baseline{21.07} & \baseline{0.81} & \baseline{294.5} & \baseline{10.81} \\
0.9999 & 0.58 & 1.35 & 8.10 & 9.34 & 28.18 & 22.24 & 0.98 & 287.7 & 11.63 \\
\end{tabular}
\end{center}
\end{table*}

\begin{table*}[h]
\begin{center}
\caption{
\textbf{Raw FD values for population size ablation} (Tabs.~\ref{tab:queue_size} and~\ref{tab:ema_beta}).
pMF-B/16 post-trained for 50 epochs with FD-Inception.
Default setting in \baselinelegend{Gray}.
$^\dagger$Statistics from current batch only.
}
\label{tab:rawfd_population}
\vspace{.3em}
\tablestyle{3pt}{1.03}
\begin{tabular}{l | c c c c c | c | c c}
setting & Incep. & ConvNeXt & DINOv2 & MAE & SigLIP & CLIP & FID & IS \\
\shline
\gc{Validation set} & \gc{1.68} & \gc{56.87} & \gc{14.19} & \gc{0.04} & \gc{0.60} & \gc{5.60} & \gc{1.68} & \gc{232.2} \\
\hline
\multicolumn{9}{l}{\hspace{-.5em} \textit{Queue size}} \\
\gc{Base} & \gc{3.31} & \gc{109.54} & \gc{143.69} & \gc{0.59} & \gc{18.77} & \gc{130.61} & \gc{3.31} & \gc{254.6} \\
\gc{0k$^\dagger$} & \gc{3.84} & \gc{226.62} & \gc{194.92} & \gc{0.72} & \gc{21.47} & \gc{168.97} & \gc{3.84} & \gc{250.9} \\
5k & 1.05 & 80.02 & 123.78 & 0.41 & 17.03 & 128.32 & 1.05 & 280.0 \\
10k & 0.93 & 77.15 & 118.50 & 0.39 & 16.98 & 128.19 & 0.93 & 283.9 \\
\baseline{50k} & \baseline{0.89} & \baseline{85.11} & \baseline{111.10} & \baseline{0.35} & \baseline{15.80} & \baseline{118.84} & \baseline{0.89} & \baseline{288.3} \\
100k & 0.93 & 91.04 & 111.22 & 0.39 & 16.11 & 118.94 & 0.93 & 288.8 \\
500k & 1.22 & 123.48 & 134.67 & 0.84 & 24.26 & 190.12 & 1.22 & 294.4 \\
\hline
\multicolumn{9}{l}{\hspace{-.5em} \textit{EMA decay rate ($\beta$)}} \\
\gc{Base} & \gc{3.31} & \gc{109.54} & \gc{143.69} & \gc{0.59} & \gc{18.77} & \gc{130.61} & \gc{3.31} & \gc{254.6} \\
\gc{0.0$^\dagger$} & \gc{3.84} & \gc{226.62} & \gc{194.92} & \gc{0.72} & \gc{21.47} & \gc{168.97} & \gc{3.84} & \gc{250.9} \\
0.9 & 0.98 & 85.05 & 117.40 & 0.37 & 15.85 & 122.42 & 0.98 & 283.6 \\
0.99 & 0.84 & 76.44 & 107.43 & 0.36 & 15.43 & 117.59 & 0.84 & 291.8 \\
\baseline{0.999} & \baseline{0.81} & \baseline{71.80} & \baseline{106.67} & \baseline{0.36} & \baseline{15.74} & \baseline{118.08} & \baseline{0.81} & \baseline{294.5} \\
0.9999 & 0.98 & 76.97 & 114.90 & 0.40 & 17.05 & 124.67 & 0.98 & 287.7 \\
\end{tabular}
\end{center}
\end{table*}

\begin{table*}[h]
\begin{center}
\caption{
\textbf{Raw FD values for representation model ablation} (Tab.~\ref{tab:backbone}).
pMF-B/16 post-trained for 50 epochs.
SIM: SigLIP+Inception+MAE.
}
\label{tab:rawfd_backbone}
\vspace{.3em}
\tablestyle{3pt}{1.03}
\begin{tabular}{l | c c c c c | c | c c}
loss & Incep. & ConvNeXt & DINOv2 & MAE & SigLIP & CLIP & FID & IS \\
\shline
\gc{Validation set} & \gc{1.68} & \gc{56.87} & \gc{14.19} & \gc{0.04} & \gc{0.60} & \gc{5.60} & \gc{1.68} & \gc{232.2} \\
\hline
Base & 3.32 & 109.75 & 143.70 & 0.59 & 18.77 & 130.59 & 3.31 & 254.6 \\
FD-Inception & 0.81 & 71.65 & 106.68 & 0.36 & 15.74 & 118.09 & 0.81 & 294.5 \\
FD-ConvNeXt & 1.64 & 19.33 & 69.94 & 0.32 & 10.51 & 110.19 & 1.64 & 281.0 \\
FD-DINOv2 & 4.88 & 121.69 & 29.93 & 0.47 & 10.23 & 88.72 & 4.89 & 347.1 \\
FD-MAE & 6.42 & 109.18 & 75.18 & 0.05 & 8.73 & 73.93 & 6.42 & 344.0 \\
FD-SigLIP & 7.72 & 152.97 & 60.57 & 0.39 & 2.18 & 60.76 & 7.71 & 399.4 \\
\hline
FD-SigLIP+Incep. & 0.89 & 54.26 & 70.80 & 0.37 & 4.13 & 77.08 & 0.89 & 307.5 \\
FD-SIM & 0.94 & 48.38 & 65.90 & 0.10 & 4.20 & 55.00 & 0.94 & 307.8 \\
\end{tabular}
\end{center}
\end{table*}

\begin{table*}[h]
\begin{center}
\caption{
\textbf{Per-representation FDr for JiT-L repurposing} (Tab.~\ref{tab:repurpose}).
JiT-L/16 post-trained for 50 epochs.
SIM: SigLIP+Inception+MAE.
}
\label{tab:detail_jit}
\vspace{.3em}
\tablestyle{3pt}{1.03}
\begin{tabular}{l | c c c c c | c | c c c}
setting & Incep. & ConvNeXt & DINOv2 & MAE & SigLIP & FDr-CLIP & FID$\downarrow$ & IS$\uparrow$ & \fdr{6}$\downarrow$ \\
\shline
\gc{JiT-L (50-step)} & \gc{1.54} & \gc{3.49} & \gc{6.10} & \gc{8.07} & \gc{19.37} & \gc{25.82} & \gc{2.59} & \gc{288.5} & \gc{10.73} \\
\gc{JiT-L (1-step)} & \gc{173.83} & \gc{142.28} & \gc{151.86} & \gc{322.61} & \gc{327.20} & \gc{170.71} & \gc{291.59} & \gc{2.0} & \gc{214.75} \\
\hline
FD-Incep. & 0.46 & 2.57 & 7.34 & 15.28 & 26.22 & 25.29 & 0.77 & 293.7 & 12.86 \\
FD-MAE & 3.89 & 3.82 & 7.87 & 2.20 & 18.21 & 19.84 & 6.52 & 280.4 & 9.30 \\
FD-SigLIP & 3.04 & 2.79 & 4.86 & 27.68 & 2.20 & 13.68 & 5.10 & 329.6 & 9.04 \\
FD-SigLIP+MAE & 2.78 & 1.91 & 4.14 & 0.96 & 2.23 & 10.97 & 4.67 & 354.0 & 3.83 \\
FD-SIM & 0.51 & 1.09 & 3.06 & 1.01 & 2.78 & 11.30 & 0.85 & 319.5 & 3.29 \\
\end{tabular}
\end{center}
\end{table*}

\begin{table*}[h]
\begin{center}
\caption{
\textbf{Raw FD values for JiT-L repurposing} (Tab.~\ref{tab:repurpose}).
JiT-L/16 post-trained for 50 epochs.
SIM: SigLIP+Inception+MAE.
}
\label{tab:rawfd_jit}
\vspace{.3em}
\tablestyle{3pt}{1.03}
\begin{tabular}{l | c c c c c | c | c c}
setting & Incep. & ConvNeXt & DINOv2 & MAE & SigLIP & CLIP & FID & IS \\
\shline
\gc{Validation set} & \gc{1.68} & \gc{56.87} & \gc{14.19} & \gc{0.04} & \gc{0.60} & \gc{5.60} & \gc{1.68} & \gc{232.2} \\
\hline
\gc{JiT-L (50-step)} & \gc{2.59} & \gc{198.19} & \gc{86.54} & \gc{0.35} & \gc{11.72} & \gc{144.69} & \gc{2.59} & \gc{288.5} \\
\gc{JiT-L (1-step)} & \gc{291.59} & \gc{8091.15} & \gc{2154.20} & \gc{13.82} & \gc{197.92} & \gc{956.77} & \gc{291.59} & \gc{2.0} \\
FD-Incep. & 0.77 & 146.26 & 104.15 & 0.65 & 15.86 & 141.76 & 0.77 & 293.7 \\
FD-MAE & 6.52 & 217.29 & 111.65 & 0.09 & 11.01 & 111.19 & 6.52 & 280.4 \\
FD-SigLIP & 5.10 & 158.87 & 68.96 & 1.19 & 1.33 & 76.69 & 5.10 & 329.6 \\
FD-SigLIP+MAE & 4.67 & 108.46 & 58.79 & 0.04 & 1.35 & 61.48 & 4.67 & 354.0 \\
FD-SIM & 0.85 & 62.03 & 43.40 & 0.04 & 1.68 & 63.34 & 0.85 & 319.5 \\
\end{tabular}
\end{center}
\end{table*}

\begin{table*}[h]
\begin{center}
\caption{
\textbf{Per-representation FDr for system-level comparison} (Tab.~\ref{tab:system}).
Our \method post-trained models in \baselinelegend{shaded} rows.
SIM: MAE+SigLIP+Incep.
}
\label{tab:detail_system}
\vspace{.3em}
\tablestyle{3pt}{1.03}
\begin{tabular}{l | c c c c c | c | c c c}
method & Incep. & ConvNeXt & DINOv2 & MAE & SigLIP & FDr-CLIP & FID$\downarrow$ & IS$\uparrow$ & \fdr{6}$\downarrow$ \\
\shline
\gc{50k validation images} & \gc{1.00} & \gc{1.00} & \gc{1.00} & \gc{1.00} & \gc{1.00} & \gc{1.00} & \gc{1.68} & \gc{232.2} & \gc{1.00} \\
\hline
\multicolumn{10}{l}{\hspace{-.5em} \textit{discrete}} \\
VAR-d30 & 1.18 & 1.70 & 5.31 & 6.83 & 11.89 & 13.31 & 1.97 & 304.6 & 6.70 \\
BAR-B~\cite{yu2026autoregressive} & 0.68 & 0.93 & 4.13 & 5.02 & 8.30 & 6.78 & 1.15 & 273.9 & 4.31 \\
BAR-L~\cite{yu2026autoregressive} & 0.61 & 0.78 & 3.29 & 4.20 & 6.60 & 5.98 & 1.01 & 281.9 & 3.57 \\
\hline
\multicolumn{10}{l}{\hspace{-.5em} \textit{latent, multi-step}} \\
\multicolumn{10}{l}{\hspace{-.2em} \textit{without semantic distillation}} \\
SiT-XL/2~\cite{ma2024sit} & 1.26 & 2.02 & 7.89 & 5.62 & 16.14 & 17.69 & 2.12 & 256.7 & 8.44 \\
MAR-L~\cite{li2025autoregressive} & 1.07 & 1.10 & 6.09 & 4.38 & 12.67 & 14.78 & 1.80 & 293.4 & 6.68 \\
FlowAR-H~\cite{ren2024flowar} & 1.00 & 1.30 & 4.68 & 4.59 & 12.75 & 12.49 & 1.68 & 274.1 & 6.13 \\
MAR-H~\cite{li2025autoregressive} & 0.93 & 0.95 & 4.95 & 3.71 & 10.07 & 13.02 & 1.56 & 299.5 & 5.61 \\
MAR-L, DeTok~\cite{yang2025latent} & 0.83 & 1.36 & 4.66 & 4.40 & 9.57 & 12.12 & 1.39 & 306.2 & 5.49 \\
\multicolumn{10}{l}{\hspace{-.2em} \textit{with semantic distillation}} \\
REG~\cite{wu2025representation} & 0.92 & 1.14 & 3.45 & 3.02 & 8.42 & 10.86 & 1.54 & 302.9 & 4.64 \\
SiT-XL/2-REPA~\cite{yu2024representation} & 0.85 & 1.22 & 4.27 & 3.85 & 9.87 & 12.65 & 1.42 & 306.1 & 5.45 \\
LightningDiT~\cite{yao2025reconstruction} & 0.85 & 1.09 & 3.76 & 3.02 & 8.47 & 10.21 & 1.42 & 294.3 & 4.57 \\
DDT-XL~\cite{wang2025ddt} & 0.75 & 1.02 & 4.26 & 4.11 & 10.16 & 13.86 & 1.26 & 309.3 & 5.70 \\
REPA-E~\cite{leng2025repa} & 0.70 & 1.28 & 2.44 & 2.52 & 5.04 & 6.28 & 1.17 & 298.3 & 3.04 \\
RAE-XL~\cite{zheng2025diffusion} & 0.69 & 1.79 & 2.11 & 3.30 & 3.79 & 7.87 & 1.16 & 261.0 & 3.26 \\
\hline
\multicolumn{10}{l}{\hspace{-.5em} \textit{latent, one-step}} \\
Drift-L~\cite{deng2026generative} & 0.91 & 2.03 & 10.35 & 6.51 & 24.12 & 21.59 & 1.53 & 257.2 & 10.92 \\
iMF-XL~\cite{geng2025improved} (1 NFE) & 1.09 & 1.72 & 7.30 & 6.09 & 17.02 & 17.14 & 1.82 & 278.9 & 8.39 \\
iMF-XL~\cite{geng2025improved} (2 NFE) & 0.96 & 1.54 & 6.31 & 5.62 & 14.91 & 15.52 & 1.61 & 289.1 & 7.48 \\
\hline
\multicolumn{10}{l}{\hspace{-.5em} \textit{pixel, multi-step}} \\
PixNerd-XL~\cite{wang2026pixnerd} & 1.25 & 1.21 & 3.57 & 3.56 & 9.12 & 11.36 & 2.10 & 318.8 & 5.01 \\
JiT-L~\cite{li2025back} & 1.54 & 3.49 & 6.10 & 8.07 & 19.37 & 25.82 & 2.59 & 288.5 & 10.73 \\
JiT-H~\cite{li2025back} & 1.18 & 2.52 & 4.28 & 5.65 & 11.91 & 20.40 & 1.97 & 296.0 & 7.66 \\
\hline
\multicolumn{10}{l}{\hspace{-.5em} \textit{pixel, one-step}} \\
Drift-L~\cite{deng2026generative} & 0.85 & 0.73 & 6.33 & 6.51 & 22.13 & 26.49 & 1.43 & 305.8 & 10.51 \\
pMF-L~\cite{lu2026one} & 1.62 & 1.36 & 6.70 & 9.72 & 20.34 & 14.81 & 2.72 & 261.7 & 9.09 \\
pMF-H~\cite{lu2026one} & 1.37 & 1.15 & 5.43 & 6.25 & 15.33 & 11.68 & 2.29 & 267.2 & 6.87 \\
\hline
\multicolumn{10}{l}{\hspace{-.5em} \textit{+ \method (ours)}} \\
\rowcolor{baselinecolor} iMF-XL, Incep. & 0.43 & 1.04 & 4.54 & 4.25 & 12.17 & 13.64 & 0.72 & 295.0 & 6.01 \\
\rowcolor{baselinecolor} iMF-XL, SIM & 0.45 & 0.69 & 2.38 & 1.02 & 3.64 & 6.54 & 0.76 & 301.3 & 2.45 \\
\rowcolor{baselinecolor} JiT-H, Incep. & 0.43 & 1.92 & 5.39 & 13.21 & 20.70 & 19.44 & 0.72 & 294.2 & 10.18 \\
\rowcolor{baselinecolor} JiT-H, SIM & 0.45 & 0.86 & 2.10 & 0.43 & 1.68 & 10.37 & 0.75 & 313.0 & 2.65 \\
\rowcolor{baselinecolor} pMF-L, Incep. & 0.44 & 0.94 & 4.60 & 4.24 & 14.34 & 12.55 & 0.73 & 293.9 & 6.19 \\
\rowcolor{baselinecolor} pMF-L, SIM & 0.47 & 0.57 & 2.21 & 0.56 & 3.03 & 5.68 & 0.78 & 309.2 & 2.09 \\
\rowcolor{baselinecolor} pMF-H, Incep. & 0.43 & 0.62 & 3.58 & 3.42 & 10.81 & 10.27 & 0.72 & 298.8 & 4.86 \\
\rowcolor{baselinecolor} pMF-H, SIM & 0.46 & 0.57 & 1.74 & 0.35 & 2.46 & 5.77 & 0.77 & 310.1 & 1.89 \\
\end{tabular}
\end{center}
\end{table*}

\begin{table*}[h]
\begin{center}
\caption{
\textbf{Raw FD values for system-level comparison} (Tab.~\ref{tab:system}).
Our \method post-trained models in \baselinelegend{shaded} rows.
}
\label{tab:rawfd_system}
\vspace{.3em}
\tablestyle{3pt}{1.03}
\begin{tabular}{l | c c c c c | c | c c}
method & Incep. & ConvNeXt & DINOv2 & MAE & SigLIP & CLIP & FID & IS \\
\shline
\gc{Validation set} & \gc{1.68} & \gc{56.87} & \gc{14.19} & \gc{0.04} & \gc{0.60} & \gc{5.60} & \gc{1.68} & \gc{232.2} \\
\hline
\multicolumn{9}{l}{\hspace{-.5em} \textit{discrete}} \\
VAR-d30 & 1.97 & 96.57 & 75.35 & 0.29 & 7.19 & 74.61 & 1.97 & 304.6 \\
BAR-B~\cite{yu2026autoregressive} & 1.15 & 52.76 & 58.61 & 0.22 & 5.02 & 38.00 & 1.15 & 273.9 \\
BAR-L~\cite{yu2026autoregressive} & 1.01 & 44.55 & 46.68 & 0.18 & 3.99 & 33.52 & 1.01 & 281.9 \\
\hline
\multicolumn{9}{l}{\hspace{-.5em} \textit{latent, multi-step}} \\
\multicolumn{9}{l}{\hspace{-.2em} \textit{without semantic distillation}} \\
SiT-XL/2~\cite{ma2024sit} & 2.12 & 114.89 & 111.86 & 0.24 & 9.76 & 99.16 & 2.12 & 256.7 \\
MAR-L~\cite{li2025autoregressive} & 1.80 & 62.33 & 86.32 & 0.19 & 7.66 & 82.81 & 1.80 & 293.4 \\
FlowAR-H~\cite{ren2024flowar} & 1.68 & 73.68 & 66.42 & 0.20 & 7.71 & 69.99 & 1.68 & 274.1 \\
MAR-H~\cite{li2025autoregressive} & 1.56 & 54.25 & 70.25 & 0.16 & 6.09 & 72.99 & 1.56 & 299.5 \\
MAR-L, DeTok~\cite{yang2025latent} & 1.39 & 77.56 & 66.14 & 0.19 & 5.79 & 67.94 & 1.39 & 306.2 \\
\multicolumn{9}{l}{\hspace{-.2em} \textit{with semantic distillation}} \\
REG~\cite{wu2025representation} & 1.54 & 64.86 & 48.93 & 0.13 & 5.09 & 60.87 & 1.54 & 302.9 \\
SiT-XL/2-REPA~\cite{yu2024representation} & 1.42 & 69.36 & 60.62 & 0.17 & 5.97 & 70.90 & 1.42 & 306.1 \\
LightningDiT~\cite{yao2025reconstruction} & 1.42 & 62.19 & 53.38 & 0.13 & 5.12 & 57.24 & 1.42 & 294.3 \\
DDT-XL~\cite{wang2025ddt} & 1.26 & 58.25 & 60.39 & 0.18 & 6.15 & 77.70 & 1.26 & 309.3 \\
REPA-E~\cite{leng2025repa} & 1.17 & 72.89 & 34.57 & 0.11 & 3.05 & 35.18 & 1.17 & 298.3 \\
RAE-XL~\cite{zheng2025diffusion} & 1.16 & 101.72 & 29.92 & 0.14 & 2.29 & 44.13 & 1.16 & 261.0 \\
\hline
\multicolumn{9}{l}{\hspace{-.5em} \textit{latent, one-step}} \\
Drift-L~\cite{deng2026generative} & 1.53 & 115.37 & 146.88 & 0.28 & 14.59 & 121.01 & 1.53 & 257.2 \\
iMF-XL~\cite{geng2025improved} (1 NFE) & 1.82 & 97.73 & 103.55 & 0.26 & 10.30 & 96.05 & 1.82 & 278.9 \\
iMF-XL~\cite{geng2025improved} (2 NFE) & 1.61 & 87.79 & 89.51 & 0.24 & 9.02 & 86.98 & 1.61 & 289.1 \\
\hline
\multicolumn{9}{l}{\hspace{-.5em} \textit{pixel, multi-step}} \\
PixNerd-XL~\cite{wang2026pixnerd} & 2.10 & 68.78 & 50.69 & 0.15 & 5.52 & 63.67 & 2.10 & 318.8 \\
JiT-L~\cite{li2025back} & 2.59 & 198.19 & 86.54 & 0.35 & 11.72 & 144.69 & 2.59 & 288.5 \\
JiT-H~\cite{li2025back} & 1.97 & 143.09 & 60.71 & 0.24 & 7.20 & 114.35 & 1.97 & 296.0 \\
\hline
\multicolumn{9}{l}{\hspace{-.5em} \textit{pixel, one-step}} \\
Drift-L~\cite{deng2026generative} & 1.43 & 41.35 & 89.84 & 0.28 & 13.39 & 148.46 & 1.43 & 305.8 \\
pMF-L~\cite{lu2026one} & 2.72 & 77.52 & 95.04 & 0.42 & 12.30 & 83.02 & 2.72 & 261.7 \\
pMF-H~\cite{lu2026one} & 2.29 & 65.45 & 76.96 & 0.27 & 9.27 & 65.47 & 2.29 & 267.2 \\
\hline
\multicolumn{9}{l}{\hspace{-.5em} \textit{+ \method (ours)}} \\
\rowcolor{baselinecolor} iMF-XL, Incep. & 0.72 & 59.13 & 64.46 & 0.18 & 7.36 & 76.46 & 0.72 & 295.0 \\
\rowcolor{baselinecolor} iMF-XL, SIM & 0.76 & 39.35 & 33.71 & 0.04 & 2.20 & 36.63 & 0.76 & 301.3 \\
\rowcolor{baselinecolor} JiT-H, Incep. & 0.72 & 109.09 & 76.51 & 0.57 & 12.52 & 108.96 & 0.72 & 294.2 \\
\rowcolor{baselinecolor} JiT-H, SIM & 0.75 & 49.09 & 29.74 & 0.02 & 1.02 & 58.10 & 0.75 & 313.0 \\
\rowcolor{baselinecolor} pMF-L, Incep. & 0.73 & 53.41 & 65.27 & 0.18 & 8.67 & 70.36 & 0.73 & 293.9 \\
\rowcolor{baselinecolor} pMF-L, SIM & 0.78 & 32.58 & 31.39 & 0.02 & 1.83 & 31.85 & 0.78 & 309.2 \\
\rowcolor{baselinecolor} pMF-H, Incep. & 0.72 & 35.47 & 50.83 & 0.15 & 6.54 & 57.58 & 0.72 & 298.8 \\
\rowcolor{baselinecolor} pMF-H, SIM & 0.77 & 32.34 & 24.69 & 0.02 & 1.49 & 32.32 & 0.77 & 310.1 \\
\end{tabular}
\end{center}
\end{table*}

\begin{table*}[h]
\begin{center}
\caption{
\textbf{Full metrics for all \method post-trained models} (expanded version of Table~\ref{tab:all_fd_models}).
Each group shows the base generator and its post-trained variants using FD-Inception and FD-SIM.
All post-trained models use 1 NFE.
\method post-trained models in \baselinelegend{shaded} rows.
SIM: SigLIP+Inception+MAE.
}
\label{tab:all_fd_models_full}
\vspace{.3em}
\tablestyle{4pt}{1.03}
\begin{tabular}{l l c c c c c c c}
method & NFE & space & \#params & \fdr{6}$\downarrow$ & FID$\downarrow$ & IS$\uparrow$ & Prec$\uparrow$ & Recall$\uparrow$ \\
\shline
\multicolumn{9}{l}{\hspace{-.5em} \textit{pMF~\cite{lu2026one} (pixel-space, one-step)}} \\
pMF-B & 1 & pixel & 118M & 13.70 & 3.31 & 254.6 & 0.81 & 0.52 \\
\rowcolor{baselinecolor} \quad + \method (Incep.) & 1 & pixel & 118M & 10.66 & 0.77 & 294.9 & 0.76 & 0.67 \\
\rowcolor{baselinecolor} \quad + \method (SIM) & 1 & pixel & 118M & 3.50 & 0.85 & 314.1 & 0.77 & 0.64 \\
pMF-L & 1 & pixel & 410M & 9.09 & 2.72 & 261.7 & 0.81 & 0.56 \\
\rowcolor{baselinecolor} \quad + \method (Incep.) & 1 & pixel & 410M & 6.19 & 0.73 & 293.9 & 0.76 & 0.68 \\
\rowcolor{baselinecolor} \quad + \method (SIM) & 1 & pixel & 410M & 2.09 & 0.78 & 309.2 & 0.76 & 0.67 \\
pMF-H & 1 & pixel & 935M & 6.87 & 2.29 & 267.2 & 0.80 & 0.59 \\
\rowcolor{baselinecolor} \quad + \method (Incep.) & 1 & pixel & 935M & 4.86 & 0.72 & 298.8 & 0.76 & 0.68 \\
\rowcolor{baselinecolor} \quad + \method (SIM) & 1 & pixel & 935M & 1.89 & 0.77 & 310.1 & 0.77 & 0.68 \\
\hline
\multicolumn{9}{l}{\hspace{-.5em} \textit{JiT~\cite{li2025back} (pixel-space, multi-step $\to$ one-step)}} \\
JiT-B & 50$\times$2$\times$2$^\dagger$ & pixel & 131M & 15.65 & 3.71 & 269.0 & 0.81 & 0.50 \\
\rowcolor{baselinecolor} \quad + \method (Incep.) & 1 & pixel & 131M & 22.48 & 0.76 & 296.4 & 0.76 & 0.67 \\
\rowcolor{baselinecolor} \quad + \method (SIM) & 1 & pixel & 131M & 5.53 & 1.00 & 325.5 & 0.78 & 0.60 \\
JiT-L & 50$\times$2$\times$2$^\dagger$ & pixel & 459M & 10.73 & 2.59 & 288.5 & 0.79 & 0.59 \\
\rowcolor{baselinecolor} \quad + \method (Incep.) & 1 & pixel & 459M & 12.75 & 0.73 & 296.6 & 0.76 & 0.67 \\
\rowcolor{baselinecolor} \quad + \method (SIM) & 1 & pixel & 459M & 3.24 & 0.77 & 317.3 & 0.77 & 0.66 \\
JiT-H & 50$\times$2$\times$2$^\dagger$ & pixel & 953M & 7.66 & 1.97 & 296.0 & 0.78 & 0.63 \\
\rowcolor{baselinecolor} \quad + \method (Incep.) & 1 & pixel & 953M & 10.18 & 0.72 & 294.2 & 0.75 & 0.68 \\
\rowcolor{baselinecolor} \quad + \method (SIM) & 1 & pixel & 953M & 2.65 & 0.75 & 313.0 & 0.76 & 0.66 \\
\hline
\multicolumn{9}{l}{\hspace{-.5em} \textit{iMF~\cite{geng2025improved} (latent-space, one-step)}} \\
iMF-B & 1 & latent & 89M & 15.29 & 3.45 & 254.2 & 0.79 & 0.53 \\
\rowcolor{baselinecolor} \quad + \method (Incep.) & 1 & latent & 89M & 11.34 & 0.79 & 296.8 & 0.76 & 0.67 \\
\rowcolor{baselinecolor} \quad + \method (SIM) & 1 & latent & 89M & 5.56 & 0.88 & 307.7 & 0.78 & 0.64 \\
iMF-L & 1 & latent & 409M & 9.06 & 1.93 & 275.1 & 0.79 & 0.61 \\
\rowcolor{baselinecolor} \quad + \method (Incep.) & 1 & latent & 409M & 6.63 & 0.75 & 293.8 & 0.76 & 0.69 \\
\rowcolor{baselinecolor} \quad + \method (SIM) & 1 & latent & 409M & 2.74 & 0.79 & 302.8 & 0.77 & 0.67 \\
iMF-XL & 1 & latent & 610M & 8.39 & 1.82 & 278.9 & 0.78 & 0.63 \\
\rowcolor{baselinecolor} \quad + \method (Incep.) & 1 & latent & 610M & 6.01 & 0.72 & 295.0 & 0.76 & 0.68 \\
\rowcolor{baselinecolor} \quad + \method (SIM) & 1 & latent & 610M & 2.45 & 0.76 & 301.3 & 0.77 & 0.67 \\
\end{tabular}
\end{center}
\end{table*}
\label{sec:appendix_detail_end}

\clearpage
\section{Text-to-Image Prompts}
\label{sec:t2i_prompts}

\input{sections/figures/fig_t2i_appendix}

\input{sections/figures/fig_t2i_extended}

Below we list the full text prompts used in Figures~\ref{fig:t2i} and~\ref{fig:t2i_extended}, in column order.

\paragraph{Main figure (Figure~\ref{fig:t2i}).}

\begin{enumerate}[leftmargin=2em, itemsep=2pt, parsep=0pt, label={\scriptsize\texttt{\#\arabic*}}]
\item A vibrant red hibiscus flower in full bloom, with its large, delicate petals spread wide open. The flower's center features a prominent stamen with a bright yellow tip, contrasting beautifully against the deep red of the petals.
\item The iconic Oriental Pearl Tower in Shanghai illuminated at night, standing tall against the dark sky. The tower is adorned with colorful lights, predominantly blue and white. In the foreground, a beautifully landscaped garden with flowers in shades of pink, purple, and white.
\item A serene coastal scene with a sandy beach in the foreground, dotted with scattered rocks. The beach leads to a rocky shoreline that meets the turquoise waters of the sea. In the background, a small, fortified structure perches atop a rocky outcrop.
\item A corner building with a classic architectural style, featuring multiple stories and a symmetrical facade. The building is painted in a light yellow hue with white decorative elements. Each window has red awnings and is adorned with potted plants.
\item A quaint, historic building with a weathered, light beige facade. The structure features multiple windows with wooden frames and balconies adorned with plants. A stone archway leads into the building. The street in front is paved with cobblestones.
\item A vibrant garden scene dominated by lush green foliage and striking red flowers. The flowers appear to be some variety of red-hot poker (Kniphofia), standing out vividly against the backdrop of glossy, dark green leaves.
\item A cozy, rustic restaurant with a warm and inviting atmosphere. The interior features wooden beams on the ceiling and polished wood flooring, giving a cabin-like feel.
\item A fishing boat navigating through choppy waters under a clear blue sky with scattered clouds. The boat is white with blue accents and features a small cabin, a deck area with railings, and various fishing equipment visible on top.
\item The exterior of a traditional-style building with a white facade and red shutters on the windows. Two flags, one on each side of the entrance, add a cultural touch. The entrance reveals an interior with wooden furniture and framed pictures.
\end{enumerate}

\paragraph{Extended figure (Figure~\ref{fig:t2i_extended}).}
\begin{enumerate}[leftmargin=2em, itemsep=2pt, parsep=0pt, label={\scriptsize\texttt{\#\arabic*}}]
\item A picturesque coastal scene with a row of colorful buildings perched on a rocky outcrop overlooking the sea. The structures are painted in vibrant hues of orange, yellow, and red.
\item A festive ceramic figurine of Santa Claus, characterized by his iconic red hat adorned with white snowflakes and a pom-pom at the tip, with a prominent white beard and mustache.
\item A sleek, modern SUV displayed at an auto show. The vehicle is light beige or silver with a shiny exterior, reflecting the bright showroom lights.
\item A charming European town square, likely in Germany, characterized by traditional half-timbered houses with red-tiled roofs and intricate wooden detailing.
\item A vintage car painted in a striking combination of light blue and white, driving on a road. The car has a classic design with whitewall tires and a rounded body shape.
\item A long, narrow indoor shopping arcade with a high, arched ceiling supported by white beams. The floor is paved with large, dark stone tiles, with rows of shops on both sides.
\item A modern, multi-story building with a unique architectural design featuring horizontal balconies that extend outward, creating a series of platforms.
\item A picturesque scene of a historic town nestled along the banks of a river, viewed through the arch of an old stone bridge, blending traditional and modern architecture.
\item A vintage red Saab 96 car, number 38, driving on a winding road surrounded by lush green trees, with its headlights on.
\item A charming, narrow cobblestone street at night, illuminated by warm, ambient lighting, flanked by traditional whitewashed buildings with blue accents on the doors.
\item A vintage rally car, an Alfa Romeo Giulia, participating in a rally event. The car is painted metallic gray with a black and white checkered hood.
\end{enumerate}
\label{sec:t2i_prompts_end}

\clearpage
\newpage


\end{document}